%% file: 0_main.tex
\documentclass{article}
\pdfoutput=1  
\usepackage{style/unites}
\usepackage{XCharter}
\usepackage[scaled=1.1]{zlmtt} 

\input{style/macros}

\newlength{\aaaicolwidth}
\newcommand{\ourmethod}{GALA}
\newcommand{\hptable}{\small\setlength{\tabcolsep}{4pt}}

\begin{document}

\makeatletter
\def\blfootnote{\gdef\@thefnmark{}\@footnotetext}
\makeatother

\makeatletter
\pagestyle{fancy}
\fancyhf{}
\renewcommand{\headrulewidth}{1pt}
\chead{\small\bf \input{1_title}
}
\cfoot{\thepage}
\thispagestyle{fancy}
\makeatother

\makeatletter
\def\icmldate#1{\gdef\@icmldate{#1}}
\icmldate{\today}
\makeatother

\makeatletter
\fancypagestyle{fancytitlepage}{
  \fancyhead{}
  \lhead{\includegraphics[height=0.8cm]{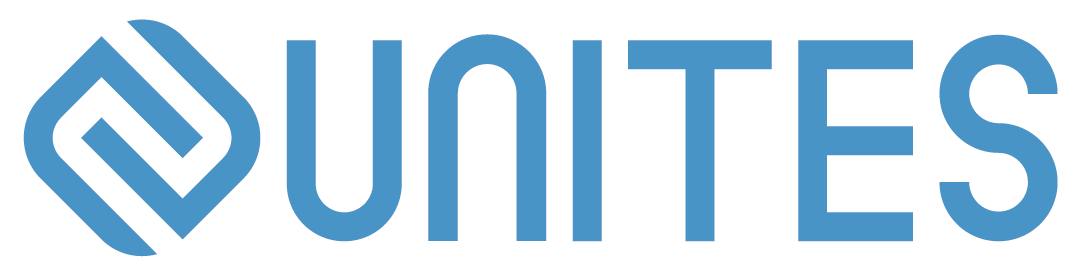}}
  \rhead{\it \@icmldate}
  \cfoot{}
}
\makeatother

\thispagestyle{fancytitlepage}

\vspace*{0.5em}

\noindent
\begin{titleblock}
    {\setlength{\parskip}{0cm}
     \raggedright
     {\setstretch{1.2}
      \LARGE\bfseries
      \input{1_title}
      \par}
    }
    \vskip 0.2cm
    
    \input{2_authors}
    \vskip 0.2cm
    
    \input{tex/0_abs}

    \vskip 0.2cm
    {\setlength{\parskip}{0cm}
     \centering
     \makebox[\linewidth]{
        \metadataformat[Code]{
            \href{https://github.com/HaochenZhang717/GALA-official-repo}{https://github.com/HaochenZhang717/GALA-official-repo}
        }
     }
    }
\end{titleblock}

\blfootnote{%
$^{\textrm{\Letter}}$ Corresponding authors: \{tianlong\}@cs.unc.edu
\\[2.5em]
\ifcsname @icmlpreprint\endcsname
  \textit{\csname @icmlpreprint\endcsname}%
\fi
}

\input{tex/1_intro}

\input{tex/3_method}

\input{tex/4_experiments}

\input{tex/2_related}

\input{tex/7_conclusion}

\input{99_acknowledgement}

\bibliography{999_reference}
\bibliographystyle{style/icml2025}

\titlespacing*{\section}{0pt}{*1}{*1}
\titlespacing*{\subsection}{0pt}{*1.25}{*1.25}
\titlespacing*{\subsubsection}{0pt}{*1.5}{*1.5}

\setlength{\abovedisplayskip}{\baselineskip} 
\setlength{\abovedisplayshortskip}{0.5\baselineskip} 
\setlength{\belowdisplayskip}{\baselineskip}
\setlength{\belowdisplayshortskip}{0.5\baselineskip}

\clearpage
\appendix
\label{sec:append}
\part*{Appendix}
{
\setlength{\parskip}{-0em}
\startcontents[sections]
\printcontents[sections]{ }{1}{}
}

\setlength{\parskip}{.35em}
\input{tex/5_appendix}

\end{document}

%% file: style/macros.tex
\usepackage[utf8]{inputenc}
\usepackage[T1]{fontenc}
\usepackage{microtype}

\usepackage{amsmath}
\usepackage{amssymb}
\usepackage{amsfonts}
\usepackage{amsthm}
\usepackage{mathtools}
\usepackage{mathrsfs}
\usepackage{physics}
\usepackage{braket}
\usepackage{slashed}
\usepackage{nicefrac}
\usepackage{textcomp}
\usepackage{dsfont}
\usepackage{bbm}
\usepackage{bm}

\usepackage{graphicx}
\usepackage{subcaption}
\usepackage[export]{adjustbox}
\usepackage{float}
\usepackage{booktabs}
\usepackage{dcolumn}
\newcolumntype{d}[1]{D{.}{.}{#1}}
\usepackage{bigstrut, tabularx, multirow, makecell, diagbox}
\usepackage{colortbl}
\usepackage{tabularray}
\UseTblrLibrary{booktabs}
\usepackage{threeparttable}
\usepackage{tablefootnote}
\usepackage{fontawesome5}

\usepackage{placeins}
\usepackage{caption}
\usepackage{footnote}
\usepackage{enumitem}
\usepackage{multicol}
\usepackage{xspace}
\usepackage{titletoc}
\usepackage{titlesec}
\usepackage[bottom]{footmisc}
\usepackage{setspace}

\usepackage{wrapfig}
\usepackage{tikz}
\usepackage{quantikz}
\usepackage{dashbox}
\usepackage{mdframed}
\usepackage{marvosym}
\usepackage{pifont}
\usepackage{CJK}
\usepackage{url}

\usepackage[table,x11names]{xcolor}
\usepackage[most]{tcolorbox}
\tcbuselibrary{breakable}
\usetikzlibrary{decorations.pathreplacing, fit}

\definecolor{primaryblue}{HTML}{0066CC}
\definecolor{accentcyan}{HTML}{00D4AA}
\definecolor{warmorange}{HTML}{FF6B35}
\definecolor{deepgray}{HTML}{2C3E50}
\definecolor{lightgray}{HTML}{F8F9FA}
\definecolor{gradientstart}{HTML}{667eea}
\definecolor{gradientend}{HTML}{764ba2}

\definecolor{citecolor}{HTML}{0071bc}
\definecolor{citeblue}{RGB}{0, 113, 188}
\definecolor{linkcolor}{HTML}{9A4D92}
\definecolor{firebrick}{rgb}{0.698,0.133,0.133}

\definecolor{paleviolet}{HTML}{E1EEFC}
\definecolor{CarolinaUltraLight}{HTML}{E7F4FC}
\definecolor{lightgrey}{RGB}{247, 247, 247}
\definecolor{shadecolor}{HTML}{EFEFEF}
\definecolor{lightyellow}{rgb}{1.0, 0.95, 0.7}
\definecolor{lightblue}{rgb}{0.90, 0.95, 1.0}
\definecolor{light-gray}{gray}{0.95}

\definecolor{darkgrey}{rgb}{0.5, 0.5, 0.5}
\definecolor{darkgreen}{rgb}{0, 0.5, 0}
\definecolor{mydarkblue}{rgb}{0,0.08,0.45}
\definecolor{mydarkblue2}{rgb}{0.133, 0.133, 0.698}
\definecolor{echodrk}{HTML}{0099cc}
\definecolor{mymauve}{rgb}{0.58,0,0.82}
\definecolor{midnightblue}{rgb}{0.1,0.1,0.44}
\definecolor{oxfordblue}{rgb}{0.0,0.13,0.28}
\definecolor{prussianblue}{rgb}{0.0,0.19,0.33}
\definecolor{coolteal}{rgb}{0, 0.45, 0.45}
\definecolor{olive}{rgb}{0.1, 0.3, 0}
\definecolor{mypurple}{rgb}{0.5,0,0.5}
\definecolor{almond}{rgb}{0.94, 0.87, 0.8}

\definecolor{blue_ampEncoding}{HTML}{DAE8FC}
\definecolor{green_encoder}{HTML}{D5E8D4}
\definecolor{purple_decoder}{HTML}{E1D5E7}
\definecolor{yellow_measure}{HTML}{FFF2CC}
\definecolor{gray_block}{HTML}{F5F5F5}
\definecolor{pink_dru}{HTML}{FAD9D5}
\definecolor{orange_v}{HTML}{FAD7AC}

\definecolor{colorA}{rgb}{1,0,0}
\definecolor{colorB}{rgb}{0,0.3,1}
\definecolor{colorC}{rgb}{0.9,0.8,0.2}
\definecolor{colorD}{rgb}{0,0.65,0}
\definecolor{lesslightgray}{rgb}{0.5,0.5,0.5}
\definecolor{fundamental}{RGB}{55, 110, 111}
\definecolor{Gred}{RGB}{219, 50, 54}
\definecolor{ToCgreen}{RGB}{0, 128, 0}
\definecolor{Sepia}{RGB}{112, 66, 20}
\definecolor{Dblue}{rgb}{0,0.08,0.45}
\definecolor{Blue}{rgb}{0, 0, 0.8}
\definecolor{blue}{rgb}{0,0,1}
\definecolor{UNCblue!10}{rgb}{0.84,0.91,0.98}
\definecolor{RowAlt}{rgb}{0.98,0.98,0.99}

\definecolor{CarolinaBlue}{HTML}{7BAFD4}        
\definecolor{CarolinaLightBlue}{HTML}{B3D4E5}   
\definecolor{CarolinaUltraLight}{HTML}{E8F4F8}  
\definecolor{CarolinaText}{HTML}{1C2B33}        

\usepackage[pagebackref=true,breaklinks=true,colorlinks,hyperfootnotes=false]{hyperref}
\hypersetup{
  colorlinks,
  citecolor=citeblue,
  linkcolor=firebrick,
  urlcolor=firebrick
}
\usepackage[nameinlink,capitalize,noabbrev]{cleveref}

\titlespacing\section{0pt}{4pt plus 4pt minus 2pt}{-2pt plus 2pt minus 2pt}
\titlespacing\subsection{0pt}{2pt plus 4pt minus 2pt}{-2pt plus 2pt minus 2pt}
\titlespacing\subsubsection{0pt}{2pt plus 4pt minus 2pt}{-2pt plus 2pt minus 2pt}

\makeatletter
\def\th@remark{%
  \thm@headfont{\bfseries}%
  \normalfont 
  \thm@preskip\topsep \divide\thm@preskip\tw@
  \thm@postskip\thm@preskip
}
\makeatother

\theoremstyle{definition}

\tcolorboxenvironment{theorem}{
  breakable,
  colback=black!10,
  colframe=white,
  width=\linewidth, 
  enlarge left by=0pt,
  enlarge right by=0pt,
  boxsep=5pt,
  boxrule=0pt,
  left=0pt,right=0pt,top=0pt,bottom=0pt,
  arc=8pt,
  before skip=\topsep,
  after skip=\topsep
}

\tcolorboxenvironment{lemma}{
  breakable,
  colback=black!10,
  colframe=white,
  width=\linewidth,
  enlarge left by=0pt,
  enlarge right by=0pt,
  boxsep=5pt,
  boxrule=0pt,
  left=0pt,right=0pt,top=0pt,bottom=0pt,
  arc=8pt,
  before skip=\topsep,
  after skip=\topsep
}

\tcolorboxenvironment{corollary}{
  breakable,
  colback=black!10,
  colframe=white,
  width=\linewidth,
  enlarge left by=0pt,
  enlarge right by=0pt,
  boxsep=5pt,
  boxrule=0pt,
  left=0pt,right=0pt,top=0pt,bottom=0pt,
  arc=8pt,
  before skip=\topsep,
  after skip=\topsep
}

\tcolorboxenvironment{proposition}{
  breakable,
  colback=black!10,
  colframe=white,
  width=\linewidth,
  enlarge left by=0pt,
  enlarge right by=0pt,
  boxsep=5pt,
  boxrule=0pt,
  left=0pt,right=0pt,top=0pt,bottom=0pt,
  arc=8pt,
  before skip=\topsep,
  after skip=\topsep
}

\tcolorboxenvironment{definition}{
  breakable,
  colback=black!10,
  colframe=white,
  width=\linewidth,
  enlarge left by=0pt,
  enlarge right by=0pt,
  boxsep=5pt,
  boxrule=0pt,
  left=0pt,right=0pt,top=0pt,bottom=0pt,
  arc=8pt,
  before skip=\topsep,
  after skip=\topsep
}

\tcolorboxenvironment{assumption}{
  breakable,
  colback=black!10,
  colframe=white,
  width=\linewidth,
  enlarge left by=0pt,
  enlarge right by=0pt,
  boxsep=5pt,
  boxrule=0pt,
  left=0pt,right=0pt,top=0pt,bottom=0pt,
  arc=8pt,
  before skip=\topsep,
  after skip=\topsep
}

\tcolorboxenvironment{claim}{
  breakable,
  colback=black!10,
  colframe=white,
  width=\linewidth,
  enlarge left by=0pt,
  enlarge right by=0pt,
  boxsep=5pt,
  boxrule=0pt,
  left=0pt,right=0pt,top=0pt,bottom=0pt,
  arc=8pt,
  before skip=\topsep,
  after skip=\topsep
}

\tcolorboxenvironment{problem}{
  breakable,
  colback=black!10,
  colframe=white,
  width=\linewidth,
  enlarge left by=0pt,
  enlarge right by=0pt,
  boxsep=5pt,
  boxrule=0pt,
  left=0pt,right=0pt,top=0pt,bottom=0pt,
  arc=8pt,
  before skip=\topsep,
  after skip=\topsep
}

\tcolorboxenvironment{question}{
  breakable,
  colback=black!10,
  colframe=white,
  width=\linewidth,
  enlarge left by=0pt,
  enlarge right by=0pt,
  boxsep=5pt,
  boxrule=0pt,
  left=0pt,right=0pt,top=0pt,bottom=0pt,
  arc=8pt,
  before skip=\topsep,
  after skip=\topsep
}

\newtcolorbox{titleblock}{
  enhanced,
  frame hidden,
  colback=CarolinaUltraLight,
  colframe=CarolinaUltraLight,
  boxrule=0pt,
  arc=10pt,
  left=14pt,
  right=14pt,
  top=14pt,
  bottom=14pt,
  width=\linewidth,
  before skip=12pt plus 4pt,
  after skip=12pt plus 4pt,
  grow to left by=1.5pt,
  grow to right by=1.5pt,
  before upper={
    \setlength{\parindent}{0cm}
    \setlength{\parskip}{0.5cm}
  }
}

\crefname{theorem}{Theorem}{Theorems}
\crefname{proposition}{Proposition}{Propositions}
\crefname{lemma}{Lemma}{Lemmas}
\crefname{corollary}{Corollary}{Corollaries}
\crefname{definition}{Definition}{Definitions}
\crefname{assumption}{Assumption}{Assumptions}
\crefname{remark}{Remark}{Remarks}
\crefname{problem}{Problem}{Problems}
\crefname{property}{Property}{property}
\crefname{question}{Question}{Questions}

\numberwithin{equation}{section}
\numberwithin{theorem}{section}
\numberwithin{proposition}{section}
\numberwithin{definition}{section}
\numberwithin{lemma}{section}
\numberwithin{assumption}{section}
\numberwithin{remark}{section}

\newcommand\metadataformat[2][]{{\small {\bfseries #1:} #2}}

\def\1{\bm{1}}

\makeatletter
\let\save@mathaccent\mathaccent
\newcommand*\if@single[3]{%
    \setbox0\hbox{${\mathaccent"0362{#1}}^H$}%
    \setbox2\hbox{${\mathaccent"0362{\kern0pt#1}}^H$}%
    \ifdim\ht0=\ht2 #3\else #2\fi
}
\newcommand*\rel@kern[1]{\kern#1\dimexpr\macc@kerna}
\newcommand*\widebar[1]{\@ifnextchar^{{\wide@bar{#1}{0}}}{\wide@bar{#1}{1}}}
\newcommand*\wide@bar[2]{\if@single{#1}{\wide@bar@{#1}{#2}{1}}{\wide@bar@{#1}{#2}{2}}}
\newcommand*\wide@bar@[3]{%
    \begingroup
    \def\mathaccent##1##2{%
        \let\mathaccent\save@mathaccent
        \if#32 \let\macc@nucleus\first@char \fi
        \setbox\z@\hbox{$\macc@style{\macc@nucleus}_{}$}%
        \setbox\tw@\hbox{$\macc@style{\macc@nucleus}{}_{}$}%
        \dimen@\wd\tw@
        \advance\dimen@-\wd\z@
        \divide\dimen@ 3
        \@tempdima\wd\tw@
        \advance\@tempdima-\scriptspace
        \divide\@tempdima 10
        \advance\dimen@-\@tempdima
        \ifdim\dimen@>\z@ \dimen@0pt\fi
        \rel@kern{0.6}\kern-\dimen@
        \if#31
        \overline{\rel@kern{-0.6}\kern\dimen@\macc@nucleus\rel@kern{0.4}\kern\dimen@}%
        \advance\dimen@0.4\dimexpr\macc@kerna
        \let\final@kern#2%
        \ifdim\dimen@<\z@ \let\final@kern1\fi
        \if\final@kern1 \kern-\dimen@\fi
        \else
        \overline{\rel@kern{-0.6}\kern\dimen@#1}%
        \fi
    }%
    \macc@depth\@ne
    \let\math@bgroup\@empty \let\math@egroup\macc@set@skewchar
    \mathsurround\z@ \frozen@everymath{\mathgroup\macc@group\relax}%
    \macc@set@skewchar\relax
    \let\mathaccentV\macc@nested@a
    \if#31
    \macc@nested@a\relax111{#1}%
    \else
    \def\gobble@till@marker##1\endmarker{}%
    \futurelet\first@char\gobble@till@marker#1\endmarker
    \ifcat\noexpand\first@char A\else
    \def\first@char{}%
    \fi
    \macc@nested@a\relax111{\first@char}%
    \fi
    \endgroup
    }
\makeatother

\DeclareMathAlphabet{\mathsfit}{\encodingdefault}{\sfdefault}{m}{sl}
\SetMathAlphabet{\mathsfit}{bold}{\encodingdefault}{\sfdefault}{bx}{n}

\let\hat\widehat



%% file: 1_title.tex
GALA: Generation-Aware Cross-Modal Alignment for Text-to-Time-Series Synthesis

%% file: 2_authors.tex
\begin{icmlauthorlist}
\mbox{Haochen Zhang$^{\,1\,}$},
\mbox{Gengwei Zhang$^{\,1\,}$},
\mbox{Laura Yao$^{\,1\,}$},
\mbox{Nicholas Konz$^{\,1\,}$}
and \mbox{Tianlong Chen$^{\,1\,\textrm{\Letter}}$}
\end{icmlauthorlist}

$^{1\,}$UNITES Lab, University of North Carolina at Chapel Hill

\{haochenz, gengweiz, lyao, nick124, tianlong\}@cs.unc.edu
 
$^{\textrm{\Letter}}$ Corresponding Author

%% file: tex/0_abs.tex
Synthesizing time series from natural language is emerging as the most expressive form of controllable time series generation. However, existing text-conditioned generators either take caption embeddings frozen from off-the-shelf text encoders, or adapt the encoder end-to-end, letting the denoising loss shape the embeddings only as a by-product. In either case, the conditioning representation is never deliberately matched to the signal modality, leaving it ill-suited to guide generation. We address this by introducing \ourmethod{}: \underline{\textbf{G}}eneration-\underline{\textbf{A}}ware cross-moda\underline{\textbf{L}} \underline{\textbf{A}}lignment for text conditional time series generation. \ourmethod{} is a two-stage approach that first contrastively couples a pretrained text encoder with a time-series foundation model into a shared embedding space with both encoders adapted to generation by an auxiliary generative loss, and then freezes the resulting caption embedding to drive a flow-matching generator. On TSFragment-600K, spanning four domains and three fragment lengths, \ourmethod{} sets a new state of the art, ranking first in $30$ of $36$ metric columns and reaching an average rank of $1.08/1.08/1.42$ at lengths $24/48/96$ against $1.92/2.00/1.75$ for the strongest baseline. We further find that generator-internal text encoders force a trade-off between fidelity and caption adherence, whereas conditioning on the aligned embedding breaks it: FID, CTTP, and JFTSD all improve at once. Ablating the auxiliary loss degrades FID, CTTP and JFTSD together, it indicates the generative term is a necessary component of the alignment rather than an add-on.

%% file: tex/1_intro.tex
\section{Introduction}

Time series generation plays an important role in domains such as energy, traffic, and healthcare, where high-quality synthetic series can support data augmentation, privacy-preserving data sharing, and simulation \citep{timegan,ttscgan}.
However, the controllability of a generator directly determines its practical value. Unconditional generation methods \citep{timevae,diffwave,diffusionts,imagentime,flowts} are not controllable. Conditioning on structured signals such as class labels or attribute vectors \citep{timevqvae,timeweaver,wavestitch,tedit} restricts the user to a predefined condition schema and supports no scenario outside it.
In contrast, a natural language description can be both flexible for users and semantically meaningful for describing time series. For example, a clinician can specify a 12-lead ECG through a clinical report \citep{diffusets}. This makes text the most promising interface for describing the morphology of time series \citep{contsg}. Consequently, text-conditional time series generation has emerged as the frontier of conditional time series generation \citep{t2s,verbalts,bridge}.

\begin{wrapfigure}{r}{\aaaicolwidth}
\centering
\includegraphics[width=\linewidth]{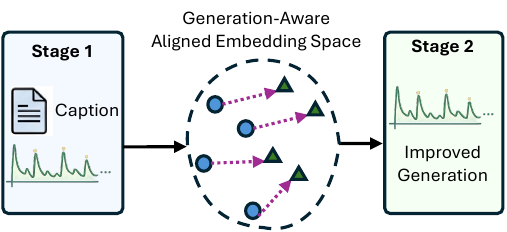}
\caption{\ourmethod{} synthesizes a time series from a free-form caption by first aligning text and series into a shared, generation-aware embedding space, then conditioning a generator on the frozen text embedding.}
\label{fig:teaser}
\end{wrapfigure}

Existing text-conditioned generators differ mainly in where generation is performed. One line of work first trains an autoencoder to compress the time series and then a generator in the latent space \citep{t2s,diffusets}, while the other line generates the raw series directly in the observation space \citep{verbalts,bridge}. Despite this difference, the two approaches share the same treatment of the text condition and lack a deliberate design of the conditioning path. T2S \citep{t2s}, DiffuSETS \citep{diffusets}, and BRIDGE \citep{bridge} obtain fixed caption representations from an off-the-shelf language embedding model and feed them to the generator, and VerbalTS \citep{verbalts} learns an adapter that reprograms the frozen text-encoder output into multi-resolution features. In all these methods, the text representation is either kept frozen from a text-only encoder or shaped only by the denoising objective, never explicitly aligned with the time-series modality.

However, explicit cross-modal alignment is effective in other time series tasks. For forecasting, TimesCLIP \citep{timesclip} contrastively aligns visual and textual views of the series, BALM-TSF \citep{balmtsf} balances the two modalities with a contrastive semantic-alignment loss, and Time-MMD \citep{timemmd} shows that aligned textual context substantially reduces forecasting error; for understanding and reasoning, TEST \citep{test} and ChatTS \citep{chatts} align series to an LLM via text prototypes or synthetic captions. Similar observations hold beyond time series \citep{clip,unclip}. Converting semantic correspondence into an explicit geometric relation improves downstream performance.

In text conditional time series generation, however, developing a good cross-modal alignment is not a direct plug-in of existing methods. We observe two challenges in our experiments. \textbf{1):} Fine-tuning the text encoders inside the generator improves caption adherence at the cost of fidelity (Table~\ref{tab:ablation-noalign}), which shows that the generation loss alone is insufficient to shape the shared latent space. \textbf{2):} Existing cross-modal alignment methods optimize the shared space for ranking and retrieval, but do not necessarily retain the fine-grained, reconstruction-relevant detail that a generator needs. We address these two challenges by learning a generation-aware text-time-series shared embedding space (Figure~\ref{fig:teaser}), which makes the text embedding faithful enough to drive the generator to better caption adherence without sacrificing fidelity.

In this work, we propose \ourmethod{} (\underline{\textbf{G}}eneration-\underline{\textbf{A}}ware cross-moda\underline{\textbf{L}} \underline{\textbf{A}}lignment), a two-stage text-to-time-series generation method. In the first stage, \ourmethod{} contrastively aligns a pretrained text encoder and a time-series foundation model into a shared embedding space, with a magnitude-aware encoding that preserves the absolute level and scale that captions often refer to. To make the alignment generation-aware, we add an auxiliary generation loss that requires the text embedding to remain sufficient to synthesize its paired series; the encoders themselves are adapted with a low-rank update. The auxiliary loss lifts caption adherence and fidelity together rather than trading them off, It is a necessary component rather than an add-on, and it works for any adaptation that reaches into the backbones, the only failing case being a trainable head on frozen features. In the second stage, the aligned text encoder is frozen and its embedding serves as the condition of a flow-matching generator. In summary, our contributions are threefold:
\begin{itemize}
\item We identify an overlooked gap in text-to-time-series generators: the conditioning representation is either frozen text-only features or adapted solely through the generation objective; in neither case is it aligned with the time-series modality, so the conditioning signal is under-optimized for generation.
\item We propose \ourmethod{}, a two-stage method that decouples cross-modal alignment from generation and makes the alignment generation-aware: a contrastive objective is combined with an auxiliary generation loss that keeps the text embedding sufficient to synthesize time series.
\item We conduct extensive experiments on TSFragment-600K across four domains and three fragment lengths, demonstrating that \ourmethod{} matches or beats state-of-the-art methods on generation fidelity while improving caption adherence by a large margin. Ablations, a probing analysis, and comparisons with alternative alignment schemes further explain where the gains come from.

\end{itemize}

%% file: tex/3_method.tex
\section{Method}

\subsection{Problem Setup}

A caption $C$ takes values in the caption space $\mathcal{C}$, and a time series $X$ takes values in the time series space $\mathcal{X} \subset \mathbb{R}^{V\times L}$, with $V$ variates and length $L$; the pair $(C,X)$ is drawn from an unknown joint distribution $\mu$ with marginals $\mu_{C}$ and $\mu_{X}$, of which the training dataset $\{(c_i,x_i)\}_{i=1}^{N}$ consists of i.i.d.\ samples. The aim of text-conditional time series generation is to learn a conditional distribution $p_{\theta}(\cdot\mid c)$, parameterized by $\theta$, that approximates $\mu(\cdot\mid c)$ for typical captions, i.e., to solve
\begin{equation}
\min_{\theta}\;
\mathbb{E}_{c\sim\mu_{C}}
\big[\,D\big(\mu(\cdot\mid c)\,\big\|\,p_{\theta}(\cdot\mid c)\big)\big],
\label{eq:objective}
\end{equation}
where $D$ is any discrepancy measure between distributions. This objective makes precise the two quality axes of our evaluation: \textbf{1)} fidelity asks that the generated series look realistic, i.e., the marginal of $p_{\theta}$ over captions matches $\mu_{X}$; \textbf{2)} caption adherence asks that each $p_{\theta}(\cdot\mid c)$ put its mass on series whose morphology matches the caption $c$, rather than collapsing to the caption-agnostic marginal.

\subsection{Technical Motivation}
Before describing \ourmethod{}, we give a short information-theoretic argument for its two design decisions: separating alignment from generation, and adding a generative term to the alignment objective. It is not a tight bound or a new theoretical result, but an account of why these two decisions are the natural ones. Anticipating the design, let $E$ denote the text encoder that maps a caption to its embedding, $U=E(C)$ that embedding, and $\mu(\cdot\mid u)$ the conditional law of $X$ given $U{=}u$ induced by $\mu$. \ourmethod{} fits $E$ in the first stage, under a contrastive loss $\mathcal{L}_{\text{align}}$ and an auxiliary generative loss $\mathcal{L}_{\text{gen}}$ that are defined below, and fits the generator in the second.

\paragraph{Explicit feature alignment as error decomposition}
In \ourmethod{}, the generator reads the caption through the text encoder: $p_\theta(\cdot\mid
c)=p_\theta(\cdot\mid E(c))$. For any such model, taking $D$ as KL divergence in
the learning objective \eqref{eq:objective} gives the exact identity
\begin{equation}
\label{eq:decomp}
\begin{aligned}
&\underbrace{\mathbb{E}_{c}[\mathrm{KL}(\mu(\cdot|c)\,\|\,p_\theta(\cdot|c))]}_{\text{end-to-end objective}} \\[2pt]
&=
\underbrace{\mathbb{E}_{c}[\mathrm{KL}(\mu(\cdot|c)\,\|\,\mu(\cdot|u))]}_{\mathcal{R}(E):\,\text{representation error}} +
\underbrace{\mathbb{E}_{u}[\mathrm{KL}(\mu(\cdot|u)\,\|\,p_\theta(\cdot|u))]}_{\text{generation error}},
\end{aligned}
\end{equation}
which follows by writing
$\log\frac{\mu(x\mid c)}{p_\theta(x\mid u)} =\log\frac{\mu(x\mid c)}{\mu(x\mid u)}+\log\frac{\mu(x\mid u)}{p_\theta(x\mid u)}$ and taking expectations; the second term reduces to $\mathbb{E}_u[\cdots]$ by the tower property, since $U=E(C)$ is
  deterministic and $X\mid U\sim\mu(\cdot\mid u)$.
Under this decomposition, $\mathcal{R}(E)$ depends only on the encoder; for any encoder, the generation term is minimized separately in Stage~2.

\paragraph{Alignment needs a generative term for generation}
Because $U=E(C)$ is a deterministic function of $C$, the alignment error is exactly a conditional mutual information, $\mathcal{R}(E)=I(X;C\mid U)=I(X;C)-I(X;U)$, where $I(X;C)$ is a constant of the
data; so minimizing $\mathcal{R}(E)$ is equivalent to maximizing $I(X;U)$, and $\mathcal{R}(E)$ reaches zero iff $U$ is a sufficient statistic of the caption for the time series. Therefore, the aim of Stage~1 is to make the caption embedding informative about its time series, not merely good at retrieval. This seemingly innocuous difference matters for text-conditional generation. The contrastive loss $\mathcal{L}_{\text{align}}$ maximizes a lower bound on $I(X;U)$ \citep{infonce,varbounds}, namely $I(X;U)\;\ge\;\log B-\mathcal{L}_{\text{align}}$; but that bound saturates at $\log B$, so once negatives are easily ranked the
objective can no longer push for more information. This yields discriminative but not necessarily generation-ready embeddings with only coarse cues about the series.

The generation-aware auxiliary loss adds pressure on the same $I(X;U)$ from the generative side. Writing the auxiliary diffusion denoiser $g_\xi$ as a conditional density estimate $q_\xi(x\mid\mathbf{u})$, its expected
log-likelihood is the variational lower bound $I(X;U)\;\ge\;H(X)+\mathbb{E}_{(x,u)}\big[\log q_\xi(x\mid\mathbf{u})\big]$, with $H(X)$ constant in $E$, and $-\mathcal{L}_{\text{gen}}$ serves as a tractable surrogate for this log-likelihood term. Unlike $\mathcal{L}_{\text{align}}$, $\mathcal{L}_{\text{gen}}$ keeps rewarding $\mathbf{u}$ for the finer information about its time series needed for reconstruction, not merely for ranking. Therefore, the auxiliary generation loss delivers better embeddings, lifting text adherence and fidelity together, provided the encoders have the capacity to be reshaped by it, which our ablation (Figure~\ref{fig:ablation-ft-dw}) confirms is necessary.

\subsection{Overview of \ourmethod{}}

\begin{figure}[t]
\centering
\includegraphics[width=\textwidth]{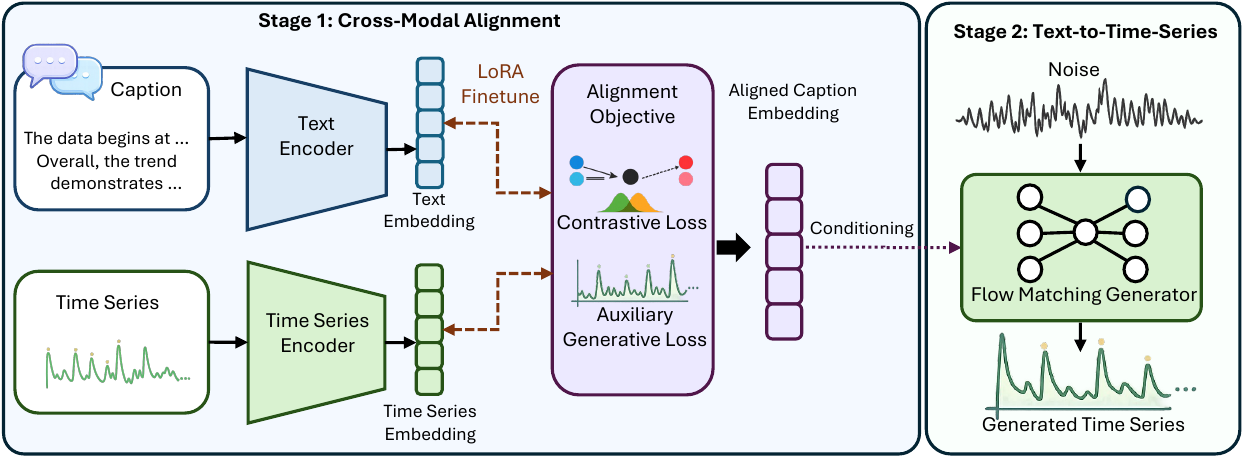}
\caption{Overview of \ourmethod{}. Stage~1 contrastively aligns a LoRA-adapted text encoder and time-series encoder into a shared space, with a generation-aware auxiliary loss keeping the text embedding sufficient to synthesize its fragment. Stage~2 freezes the aligned text encoder and trains a flow-matching generator on its embedding.}
\label{fig:overview}
\end{figure}

\ourmethod{} generates a time-series fragment from a free-form natural-language description in two decoupled stages (Figure~\ref{fig:overview}). Stage~1 learns a generation-aware shared embedding space in which a time series and its caption are mapped next to one another, by adapting a pretrained text encoder and time-series foundation model. Stage~2 freezes the aligned text encoder and trains a conditional flow-matching generator that turns the text embedding into a fragment. This decoupling is the core design: rather than learning the conditioning representation inside the generator under the reconstruction objective alone, the dedicated alignment stage produces semantically structured conditioning signal for Stage~2 to reuse.

\subsection{Stage 1: Cross-Modal Alignment}
We build a dual encoder \citep{clip} to map captions and time series onto a shared $d$-dimensional unit hypersphere.
Captions are encoded with Embedding-Gemma-300M \citep{embeddinggemma}. The backbone is kept frozen and adapted with LoRA \citep{lora} on the attention projections $\{q,k,v,o\}$. A caption $c$ is tokenized and passed through the encoder; we mean-pool the last hidden states over non-padding tokens, apply a linear projection $W_t$, and $\ell_2$-normalize, $\mathbf{u} = W_t\,\mathrm{pool}(G_\phi(c)) \big/\lVert W_t\,\mathrm{pool}(G_\phi(c))\rVert_2 \in \mathbb{R}^{d}$, where $G_\phi$ denotes the LoRA-adapted text encoder.
Time series are encoded with Chronos-2 \citep{chronos2}, an encoder-only
time-series foundation model, again frozen and LoRA-adapted on its
attention projections. One additional change is essential: Chronos-2's instance normalization re-centers and re-scales every series individually, discarding the absolute level and scale that captions often refer to; we replace it with a dataset-level global $z$-score normalization so that magnitude information survives into the embedding (Table~\ref{tab:norm}). The encoded fragment is read out at its [Reg] token at the second-to-last encoder
position, averaged over variates, projected by $W_s$, and
$\ell_2$-normalized, $\mathbf{w} = W_s\,\mathrm{h}_{\mathrm{glob}}(C_\psi(x)) \big/ \lVert W_s\,\mathrm{h}_{\mathrm{glob}}(C_\psi(x))\rVert_2 \in \mathbb{R}^{d}$, with $C_\psi$ the LoRA-adapted Chronos-2 encoder.

We optimize a contrastive loss plus an auxiliary generation loss to make the two modalities' features aligned and generation-oriented.
\textbf{1) Contrastive objective:}
For a batch of $B$ caption--fragment pairs we form the scaled
cosine-similarity matrix
$S_{ij} = \mathbf{u}_i^\top \mathbf{w}_j / \tau$ with a fixed temperature
$\tau = 0.07$. Each caption is paired with a single fragment,
so the positives lie on the diagonal and we minimize the standard
symmetric InfoNCE loss \citep{infonce}, $\mathcal{L}_{\text{t}\to\text{s}} = -\frac{1}{B}\sum_{i=1}^{B} \log \frac{\exp(S_{ii})}{\sum_{k=1}^{B}\exp(S_{ik})}$,
$\mathcal{L}_{\text{s}\to\text{t}} = -\frac{1}{B}\sum_{i=1}^{B} \log \frac{\exp(S_{ii})}{\sum_{k=1}^{B}\exp(S_{ki})}$,
$\mathcal{L}_{\text{align}} = \tfrac{1}{2}\big(\mathcal{L}_{\text{t}\to\text{s}} + \mathcal{L}_{\text{s}\to\text{t}}\big)$.
\textbf{2) Generation-aware auxiliary loss:}
Though contrastive alignment makes the text embedding discriminative for time series, it does not guarantee that the embedding retains enough information to steer the generator. We add an auxiliary objective to demand exactly this. A lightweight conditional denoiser $g_\xi$, conditioned only on the text embedding $\mathbf{u}$, is trained to
synthesize the paired time series under the same flow-matching loss used in
Stage~2; writing it as $\mathcal{L}_{\text{gen}}(g_\xi;\mathbf{u},x)$,
the Stage-1 objective is $\mathcal{L}_{\text{stage-1}}
= \mathcal{L}_{\text{align}}
  + \lambda_{\text{dw}}\,
    \mathbb{E}_{(c,x)}[\mathcal{L}_{\text{gen}}(g_\xi;\mathbf{u},x)]$.
The denoiser is discarded after Stage~1; its only purpose is
to shape the embedding. Crucially, it is conditioned on the \emph{text}
side, so a low loss is only achievable when the caption embedding encodes
the fragment's morphology well enough to drive synthesis.

\subsection{Stage 2: Conditional Generation}
Stage~2 trains a conditional generator on top of the frozen aligned text encoder, with the caption embedding $\mathbf{u}$ as the only conditioning signal. We keep the generator in the observation domain. A time series $x\in\mathbb{R}^{V \times L}$ is mapped to a square image $X=\mathcal{D}(x)\in\mathbb{R}^{V\times m\times m}$ by a delay embedding \citep{imagentime}, whose $i$-th column is the window $x[i\cdot s : i\cdot s + m]$ (embedding dimension $m$, stride $s$); $\mathcal{D}$ is invertible, so a generated image is read back to a series by $\mathcal{D}^{-1}$. For $f_\theta$ we adopt a DiT-style diffusion transformer \citep{dit} and train it under the JiT recipe \citep{jit}, in which the network outputs a clean-signal estimate while the loss is imposed on velocities. Given $X$, we draw $\epsilon\sim\mathcal{N}(0,\sigma^2 I)$ together with a logit-normal timestep $t\in(0,1)$ and interpolate the two linearly, $z_t = t\,X+(1-t)\,\epsilon$; the velocity that carries $z_t$ to the data is $v^\star=(X-z_t)/(1-t)$. Feeding $z_t$, $t$ and $\mathbf{u}$ to the network gives $\hat X=f_\theta(z_t,t,\mathbf{u})$, whose induced velocity $\hat v=(\hat X-z_t)/(1-t)$ is matched to $v^\star$ by minimizing
$\mathcal{L}_{\text{gen}} = \mathbb{E}[\lVert v^\star-\hat v\rVert_2^2]$.
Sampling runs the process backwards: a 50-step deterministic Heun solver transports Gaussian noise along the learned velocity field, and $\mathcal{D}^{-1}$ maps the resulting image back to a time series.


%% file: tex/4_experiments.tex
\section{Experiments}


\subsection{Experimental Setup and Implementation Details}
We use \mbox{TSFragment-600K} \citep{t2s}, a large-scale benchmark ($>$600K pairs) purpose-built for text-to-time-series generation, on its four sources (ETTh1s, ETTm1, electricity, traffic) at three lengths ($24$, $48$, $96$), each split $90/10$ into training and held-out test sets.
We compare against VerbalTS \citep{verbalts}, T2S \citep{t2s}, DiffuSETS \citep{diffusets}, BRIDGE \citep{bridge} and Text2Motion \citep{text2motion}, all run from their official implementations with adaptation for text-conditional generation.
Fidelity is measured by FID and caption adherence by CTTP and JFTSD \citep{contsg}, all three scored in the embedding space of an evaluation aligner following the protocol of \citet{contsg}.
Its text tower is BGE-large-en-v1.5 \citep{bge} and its time series tower is a convolutional encoder trained from scratch, so the evaluator shares no weights with either \ourmethod{} tower and inherits nothing from Chronos-2.
For the caption-generation probe, we adopt the CaTS-Bench metrics \citep{cats}.
All experiments run on a single NVIDIA RTX~PRO 6000 Blackwell GPU; architectures, hyperparameters and code are in the supplements.

\subsection{Main Results}

\begin{table}[t]
\centering
{
\setlength{\tabcolsep}{1mm}
\resizebox{\linewidth}{!}{%
\begin{tabular}{lccccccccccccc}
\toprule
\textbf{Method} & \multicolumn{3}{c}{\textbf{ETTh1s}} & \multicolumn{3}{c}{\textbf{ETTm1}} & \multicolumn{3}{c}{\textbf{Electricity}} & \multicolumn{3}{c}{\textbf{Traffic}} & \textbf{Avg.} \\

\cmidrule(lr){2-4} \cmidrule(lr){5-7} \cmidrule(lr){8-10} \cmidrule(lr){11-13} \cmidrule(lr){14-14}
 & FID$\downarrow$ & CTTP$\uparrow$ & JFTSD$\downarrow$ & FID$\downarrow$ & CTTP$\uparrow$ & JFTSD$\downarrow$ & FID$\downarrow$ & CTTP$\uparrow$ & JFTSD$\downarrow$ & FID$\downarrow$ & CTTP$\uparrow$ & JFTSD$\downarrow$ & Rank$\downarrow$ \\

\midrule
\multicolumn{14}{l}{\emph{Length }$L=24$} \\
\midrule
DiffuSETS & 0.089 & 0.05 & 0.56 & 0.877 & -0.04 & 1.13 & 0.102 & 0.10 & 0.31 & 0.080 & 0.14 & 0.23 & 5.08 \\
T2S & 0.078 & 0.07 & 0.54 & 0.254 & 0.11 & 0.53 & 0.166 & 0.09 & 0.37 & 0.403 & 0.07 & 0.56 & 5.17 \\
BRIDGE & 0.028 & 0.03 & 0.53 & 0.050 & 0.06 & 0.38 & 0.057 & 0.07 & 0.29 & 0.117 & 0.05 & 0.33 & 4.67 \\
Text2Motion & 0.067 & 0.22 & 0.38 & 0.025 & 0.36 & 0.15 & 0.055 & 0.24 & 0.18 & 0.077 & 0.26 & 0.17 & 3.08 \\
VerbalTS & \textbf{0.019} & \underline{0.43} & \underline{0.21} & \underline{0.019} & \underline{0.52} & \underline{0.08} & \underline{0.025} & \underline{0.40} & \underline{0.09} & \underline{0.073} & \underline{0.35} & \underline{0.14} & \underline{1.92} \\
\textbf{\ourmethod{}} & \underline{0.024} & \textbf{0.71} & \textbf{0.07} & \textbf{0.015} & \textbf{0.64} & \textbf{0.03} & \textbf{0.018} & \textbf{0.52} & \textbf{0.04} & \textbf{0.050} & \textbf{0.45} & \textbf{0.09} & \textbf{1.08} \\
\midrule
\multicolumn{14}{l}{\emph{Length }$L=48$} \\
\midrule
DiffuSETS & 0.054 & 0.04 & 0.53 & 0.131 & 0.09 & 0.38 & 0.063 & 0.09 & 0.22 & 0.152 & 0.04 & 0.33 & 4.83 \\
T2S & 0.098 & 0.06 & 0.56 & 0.177 & 0.12 & 0.41 & 0.192 & 0.11 & 0.34 & 0.489 & 0.08 & 0.61 & 5.33 \\
BRIDGE & 0.035 & 0.03 & 0.52 & 0.037 & 0.06 & 0.32 & \underline{0.031} & 0.05 & 0.23 & 0.097 & 0.05 & 0.28 & 4.33 \\
Text2Motion & 0.094 & 0.16 & 0.46 & 0.042 & 0.30 & 0.16 & 0.068 & 0.22 & 0.16 & 0.089 & 0.25 & 0.16 & 3.42 \\
VerbalTS & \underline{0.029} & \underline{0.43} & \underline{0.22} & \underline{0.022} & \underline{0.46} & \underline{0.08} & 0.038 & \underline{0.40} & \underline{0.08} & \textbf{0.062} & \underline{0.38} & \underline{0.10} & \underline{2.00} \\
\textbf{\ourmethod{}} & \textbf{0.022} & \textbf{0.64} & \textbf{0.08} & \textbf{0.011} & \textbf{0.58} & \textbf{0.03} & \textbf{0.020} & \textbf{0.46} & \textbf{0.04} & \underline{0.069} & \textbf{0.48} & \textbf{0.09} & \textbf{1.08} \\
\midrule
\multicolumn{14}{l}{\emph{Length }$L=96$} \\
\midrule
DiffuSETS & 1.187 & 0.00 & 1.41 & 1.410 & 0.01 & 1.52 & 1.017 & 0.00 & 1.13 & 1.258 & 0.02 & 1.35 & 6.00 \\
T2S & 0.171 & 0.06 & 0.60 & 0.183 & 0.13 & 0.37 & 0.526 & 0.06 & 0.65 & 0.606 & 0.10 & 0.72 & 4.67 \\
BRIDGE & 0.040 & 0.02 & 0.52 & 0.033 & 0.06 & 0.27 & \underline{0.046} & 0.04 & 0.22 & \underline{0.073} & 0.09 & 0.25 & 3.67 \\
Text2Motion & 0.102 & 0.20 & 0.41 & 0.171 & 0.24 & 0.28 & 0.171 & 0.18 & 0.26 & 0.122 & 0.25 & 0.19 & 3.50 \\
VerbalTS & \textbf{0.024} & \underline{0.46} & \underline{0.17} & \textbf{0.023} & \underline{0.45} & \underline{0.06} & 0.053 & \underline{0.36} & \underline{0.09} & \textbf{0.057} & \underline{0.42} & \textbf{0.08} & \underline{1.75} \\
\textbf{\ourmethod{}} & \underline{0.024} & \textbf{0.65} & \textbf{0.07} & \underline{0.024} & \textbf{0.51} & \textbf{0.04} & \textbf{0.032} & \textbf{0.40} & \textbf{0.06} & 0.088 & \textbf{0.47} & \underline{0.11} & \textbf{1.42} \\
\bottomrule
\end{tabular}}}
\caption{Comparison with existing text-to-time-series generation methods on TSFragment-600K across four domains and three lengths ($L\in\{24,48,96\}$). Each entry is the mean over $3$ independent sampling seeds. Per-cell standard deviations are omitted for space and are uniformly small: the largest of all $216$ is $0.0165$ (DiffuSETS, FID on ETTh1s-$96$), the largest of \ourmethod{}'s $36$ is $0.0033$, and $98\%$ are below $0.01$. Significance test is deferred to Technical Supplement. Bold and underline mark the best and second-best per column, ties broken by the unrounded means. Avg. Rank is the mean rank over that length's $12$ columns.}
\label{tab:main-comparison}
\end{table}

\paragraph{\ourmethod{} achieves new SoTA.}
Table~\ref{tab:main-comparison} compares our method against five existing methods.
Our model is best in $30$ out of the
$36$ columns, for a near-perfect average rank of $1.08$/$1.08$/$1.42$ at
$L=24/48/96$ versus $1.92$/$2.00$/$1.75$ for the strongest baseline
(VerbalTS) and $\geq3.0$ for all others. This observation is consistent with the benchmark study of \citet{contsg}, which also identifies VerbalTS as the strongest existing baseline.
The advantage is largest on caption adherence: we obtain the highest CTTP in every one of the $12$ dataset--length settings, and also the lowest JFTSD in $11$ of $12$.
On fidelity we remain best on $7$ of $12$ FID columns and competitive elsewhere.
Traffic accounts for half of the six columns we do not lead, and is also by far the least diverse source; see the technical supplement.
In short, our method matches or beats prior work on fidelity and improves caption adherence by a large margin.

\paragraph{Not all alignment transfers to generation.}

\begin{figure}[!tbp]
\centering
\begin{minipage}[t]{0.48\textwidth}\vspace{0pt}
\centering
\includegraphics[width=\linewidth]{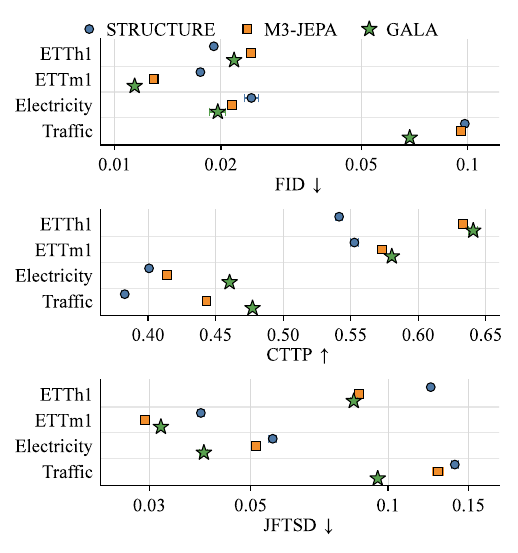}
\caption{Stage-1 alignment objectives on the four TSFragment-600K domains at length $48$, swapping only the alignment stage: STRUCTURE \citep{structure}, M3-JEPA \citep{m3jepa}, and \ourmethod{}. Markers are the mean over $3$ sampling seeds and error bars the sample std, often narrower than the marker. FID and JFTSD use a logarithmic $x$-axis.}
\label{fig:align-compare}
\end{minipage}
\hfill
\begin{minipage}[t]{0.48\textwidth}\vspace{0pt}
\centering
\adjustbox{max width=\linewidth}{\begin{tabular}{lccc}
\toprule
\textbf{Conditioning} & \textbf{FID}\,$\downarrow$ & \textbf{CTTP}\,$\uparrow$ & \textbf{JFTSD}\,$\downarrow$ \\
\midrule
No-Align (frozen) & $0.0303$ & $0.3169$ & $0.2784$ \\
No-Align (LoRA)   & $0.0341$ & $0.4219$ & $0.2207$ \\
Align (one-stage) & $0.0274$ & $0.6118$ & $0.1098$ \\
\midrule
\textbf{Align (two-stage)} & $\mathbf{0.0218}$ & $\mathbf{0.6407}$ & $\mathbf{0.0841}$ \\
\bottomrule
\end{tabular}}
\captionof{table}{Effect of the feature-alignment stage on ETTh1s-48. No-Align drops Stage-1 and conditions the generator directly on the frozen or LoRA-tuned text encoder; Align (one-stage) trains aligner and generator in a single run; Align (two-stage) is \ourmethod{}. Mean over $3$ sampling seeds; bold marks the best per column}
\label{tab:ablation-noalign}
\end{minipage}
\end{figure}

We isolate how Stage~1 shapes the conditioning embedding by swapping in two representative alignment methods with other factors fixed.
STRUCTURE \citep{structure} freezes both backbones and trains a per-side head, adding a structure-preserving regularizer that ties each aligned feature to its frozen original's geometry.
M3-JEPA \citep{m3jepa} regresses one modality's latent from the other with a Mixture-of-Experts predictor, plus an InfoNCE term.
Both are purely representational.
\ourmethod{} instead makes the embedding the sole condition of a denoiser that must synthesize the fragment, optimizing it for generative sufficiency.
Figure~\ref{fig:align-compare} shows the results.
On fidelity, \ourmethod{} obtains the lowest FID on three of the four domains.
For caption adherence: \ourmethod{} attains the highest CTTP on every domain, and the lowest JFTSD on three of four.
M3-JEPA beats STRUCTURE on CTTP and JFTSD across all four domains.
This shows that objectives forcing the embedding to encode series morphology better enable the generator to follow the caption.

\subsection{Ablation Study}

\paragraph{Dedicated alignment stage leads to better generation.}
We test the central claim of \ourmethod{}: a dedicated alignment stage produces a better conditioning signal. The No-Align variants condition the generator on the text encoder directly, frozen or LoRA-tuned; in both, the mean-pooling and linear projector on top are still trained jointly with the generator, so the conditioning representation is shaped only by the generation objective. Align (one-stage) instead trains the full Stage-1 aligner, contrastive loss included, together with the generator. Two observations follow from Table~\ref{tab:ablation-noalign}: \textbf{1)} against No-Align (frozen), No-Align (LoRA) improves caption adherence but sacrifices fidelity, so without an alignment stage the two quality axes trade off; \textbf{2)} Align (one-stage) beats both No-Align variants but trails Align (two-stage), so alignment pays off fully only as a separate stage.

\begin{figure}[!tbp]
\centering
\begin{minipage}[t]{0.48\textwidth}\vspace{0pt}
\centering
\includegraphics[width=\linewidth]{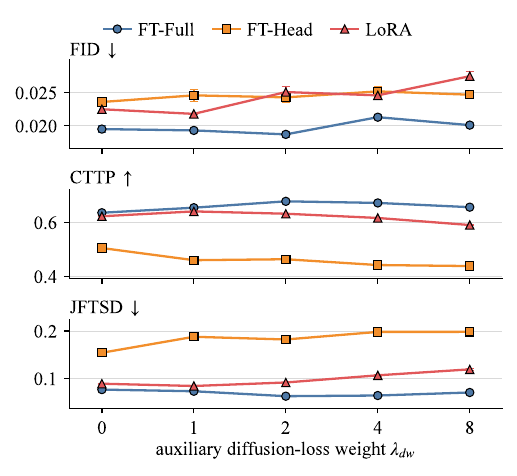}
\caption{Aligner design sweep on ETTh1s-48: fine-tuning method (FT-Full\,/\,FT-Head\,/\,LoRA) against the auxiliary-loss weight $\lambda_{\text{dw}}\in\{0,1,2,4,8\}$ (x-axis, evenly spaced). Markers are the mean over $3$ sampling seeds and bars the sample std; LoRA\,/\,$\lambda_{\text{dw}}{=}1$ is our default.}
\label{fig:ablation-ft-dw}
\end{minipage}
\hfill
\begin{minipage}[t]{0.48\textwidth}\vspace{0pt}
\centering
\includegraphics[width=\linewidth]{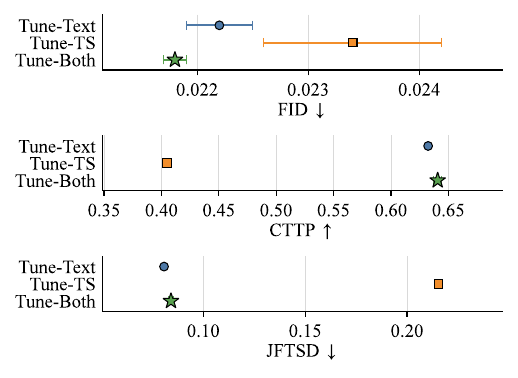}
\caption{Alignment-side ablation on ETTh1s-48: LoRA-tuning the text tower only, the series tower only, or both (our default) in Stage~1, each frozen and scored with the identical Stage-2 generator. Markers are the mean over $3$ sampling seeds; bars are the sample std.}
\label{fig:ablation-tuneside}
\end{minipage}
\end{figure}

\paragraph{The auxiliary loss is necessary in Stage~1.}
We sweep how the encoders are adapted (LoRA, FT-Full with both backbones trainable, or FT-Head with a trainable head on frozen backbones) against the auxiliary-loss weight $\lambda_{\text{dw}}\in\{0,1,2,4,8\}$ (Figure~\ref{fig:ablation-ft-dw}), where $\lambda_{\text{dw}}{=}0$ leaves the aligner purely contrastive.
The loss is what produces the gain. At its best weight, each architecture that adapts the backbone improves all three metrics at once over $\lambda_{\text{dw}}{=}0$: by $3.1$/$2.9$/$5.4\%$ (FID/CTTP/JFTSD) for LoRA at $\lambda_{\text{dw}}{=}1$ and $4.1$/$6.6$/$18.2\%$ for FT-Full at $\lambda_{\text{dw}}{=}2$, the caption-adherence gains spanning $4$ to $25$ sampling standard deviations. FT-Head shows that an adaptable backbone is required, since a frozen one leaves the loss nothing to reshape and every non-zero weight is worse than $\lambda_{\text{dw}}{=}0$.
The sweep also measures sensitivity to $\lambda_{\text{dw}}$. Each adapting architecture has an interior optimum, at $\lambda_{\text{dw}}{=}1$ for LoRA and $2$ for FT-Full, and raising the weight further gives part of the gain back. We adopt LoRA at $\lambda_{\text{dw}}{=}1$, which trains only a small low-rank update instead of both backbones in full.

\paragraph{Tuning the text encoder is what really matters.}
We LoRA-tune the text tower only (Tune-Text), the series tower only (Tune-TS), or both (Tune-Both, our default) in Stage~1; on an un-tuned tower the backbone stays frozen and only its single linear projector to the shared space is trained. Figure~\ref{fig:ablation-tuneside} shows that tuning the text encoder is what really matters: leaving it frozen collapses caption adherence, with Tune-TS at CTTP $0.405$ against $0.632$ for Tune-Text and $0.641$ for Tune-Both, and JFTSD $0.215$ against $0.081$ and $0.084$. Fidelity barely moves, FID spanning only $0.0218$--$0.0234$ across the three variants. Adding the series tower on top of the text tower buys a further gain on CTTP ($0.641$ vs.\ $0.632$) and FID while costing a little JFTSD, and we adopt Tune-Both as the default.

\paragraph{Magnitude-preserving normalization carries caption adherence.}
Chronos-2 normalizes every fragment individually, erasing the absolute level and scale that captions frequently name; Stage~1 replaces this with a dataset-level global $z$-score. We ablate the choice by fitting the aligner under each normalization and training the identical Stage-2 generator on top, on all four domains at length $48$. Table~\ref{tab:norm} shows the effect is concentrated on caption adherence: the global $z$-score wins CTTP and JFTSD on every domain, by as much as $0.22$ CTTP on ETTm1, whereas FID is close and split ($3$ of $4$ domains in our favour, ETTh1s excepted). Preserving magnitude therefore buys caption adherence at essentially no cost in fidelity.

\begin{table}[!tbp]
\centering
\begin{minipage}[t]{0.48\textwidth}\vspace{0pt}
\centering
{\small
\setlength{\tabcolsep}{4pt}
\adjustbox{max width=\linewidth}{\begin{tabular}{llcccc}
\toprule
\textbf{Metric} & \textbf{Norm.} & \textbf{ETTh1s} & \textbf{ETTm1} & \textbf{Elec.} & \textbf{Traffic} \\
\midrule
FID\,$\downarrow$ & global & $0.022$ & $\mathbf{0.011}$ & $\mathbf{0.020}$ & $\mathbf{0.069}$ \\
 & per-sample & $\mathbf{0.018}$ & $0.018$ & $0.024$ & $0.087$ \\
\addlinespace
CTTP\,$\uparrow$ & global & $\mathbf{0.641}$ & $\mathbf{0.580}$ & $\mathbf{0.460}$ & $\mathbf{0.477}$ \\
 & per-sample & $0.546$ & $0.358$ & $0.431$ & $0.432$ \\
\addlinespace
JFTSD\,$\downarrow$ & global & $\mathbf{0.084}$ & $\mathbf{0.032}$ & $\mathbf{0.040}$ & $\mathbf{0.095}$ \\
 & per-sample & $0.128$ & $0.135$ & $0.054$ & $0.122$ \\
\bottomrule
\end{tabular}}}
\caption{Time-series normalization in the Stage-1 aligner, on all four domains at length $48$. global is our dataset-level global $z$-score, which preserves absolute level and scale; per-sample is Chronos-2's native instance normalization. Each aligner is frozen and scored by training the identical Stage-2 generator on top. Mean over $3$ sampling seeds; bold marks the better normalization per domain and metric.}
\label{tab:norm}
\end{minipage}
\hfill
\begin{minipage}[t]{0.48\textwidth}\vspace{0pt}
\centering
\adjustbox{max width=\linewidth}{\begin{tabular}{lccc}
\toprule
\textbf{Metric ($\uparrow$)} & \textbf{Aligned } & \textbf{Chronos-2} & \textbf{TSLM} \\
\midrule
\multicolumn{4}{l}{\emph{Text similarity}}\\
BERT-F1       & \textbf{0.737} & 0.735 & 0.736 \\
SimCSE        & \textbf{0.904} & 0.889 & 0.893 \\
BLEU          & 0.171 & 0.169 & \textbf{0.175} \\
ROUGE-L       & \textbf{0.407} & 0.398 & 0.403 \\
METEOR        & 0.391 & 0.388 & \textbf{0.395} \\
\addlinespace
\multicolumn{4}{l}{\emph{Numeric \& statistical inference}}\\
Num.\ Fidelity & \textbf{0.471} & 0.302 & 0.360 \\
min-acc        & \textbf{0.545} & 0.097 & 0.178 \\
max-acc        & \textbf{0.772} & 0.151 & 0.228 \\
\bottomrule
\end{tabular}}
\caption{Probing how much textual information the encoded series feature retains, averaged over the four datasets at length $48$; bold marks the best per row. Aligned is our cross-modally aligned feature; the pretrained Chronos-2~\citep{chronos2} variant uses the same global $z$-score normalization, differing only in alignment; TSLM~\citep{tslm} is a further encoder baseline.}
\label{tab:caption-eval}
\end{minipage}
\end{table}

\paragraph{The aligned feature better recovers its caption. \label{sec:probe_align_feature}}

Our method targets generation, but the alignment stage also yields a series encoder whose features should carry text-relevant information. As an auxiliary probe, for each encoder variant we LoRA-fine-tune an LLM (Qwen2.5-7B-Instruct) to caption a fragment from its encoded feature and score the captions against the references with CaTS-Bench metrics~\citep{cats} (Table~\ref{tab:caption-eval}). To isolate alignment from the magnitude-aware encoding, the Chronos-2 variant is given the same global $z$-score normalization, so the two differ only in whether Stage-1 alignment is applied. On the text-overlap metrics all three variants are close, so their coarse semantic content is comparable; they separate on numeric fidelity and min/max recovery, where the aligned feature is clearly best. Since the un-aligned Chronos-2 shares the magnitude-preserving normalization, this gain is attributable to alignment rather than normalization. We read this modestly: alignment does not erase text-relevant information and appears to retain level- and range-related cues, but we do not claim it is an optimal captioning feature.

%% file: tex/2_related.tex
\section{Related Work}

\subsection{Time Series Generation}
Research on time series generation has progressed from unconditional synthesis \citep{timegan,timevae,diffwave,diffusionts,imagentime,flowts,imagenfew}, through conditioning on structured, machine-readable signals \citep{contsg}, to conditioning on free-form text. Structured control attaches either a discrete class (TTS-CGAN \citep{ttscgan}, TimeVQVAE \citep{timevqvae}) or low-dimensional attribute metadata (Time Weaver \citep{timeweaver}, WaveStitch \citep{wavestitch}, TEdit \citep{tedit}) to the generator; it is effective but bounded by a predefined schema.

Free-form text enables flexible conditioning \citep{contsg}. Recent generators include DiffuSETS \citep{diffusets} for ECG from clinical reports, T2S \citep{t2s}, a latent diffusion model with a length-adaptive autoencoder, VerbalTS \citep{verbalts}, with multi-view denoising for tighter text–signal alignment, and BRIDGE \citep{bridge}, with multi-agent optimization of text control. In all of them the conditioning path is either a frozen text-only encoder or trained solely under the reconstruction objective, so the representation is never aligned with the series modality. We instead decouple alignment from generation: a dedicated contrastive stage produces a text embedding aligned with the series domain, then frozen and reused by the generator.

\subsection{Multimodal Feature Alignment}

Feature alignment maps heterogeneous modalities into a shared representation, making semantic correspondence a geometric relation; the technique is typically chosen for the downstream task it serves. Contrastive dual-encoder alignment \citep{clip,align,coca} delivers strong zero-shot classification and cross-modal retrieval, while connector-based methods align a modality to a frozen language model for open-ended understanding, as in Flamingo \citep{flamingo} and LLaVA \citep{llava}.

Multimodal alignment has also been applied to time series, chiefly with language models \citep{tsxmodalsurvey}: Time-LLM \citep{timellm} and TimesCLIP \citep{timesclip} align series to an LLM representation space for forecasting, and TEST \citep{test} and ChatTS \citep{chatts} do so via text prototypes or synthetic captions for question answering and explanation. Our setting is instead generation, where the alignment supplies the conditioning signal.

%% file: tex/7_conclusion.tex
\section{Conclusion}

We revisited text-to-time-series generation from the standpoint of
representation quality and pinpointed a blind spot: by taking the conditioning signal
from a text-only encoder or shaping it through the generative loss alone, current
methods leave the language--signal correspondence implicit.
\ourmethod{} remedies this with a dedicated alignment stage held apart from synthesis. An information-theoretic reading clarifies why the split is reasonable (the objective factors into a representation and a generation term) and why a contrastive criterion alone falls short (too weak once negatives are easy to rank). Restoring the lost guidance with a generation-aware auxiliary loss, applied to encoders that are themselves adapted, lets both quality facets climb together. Experiments across four domains and three lengths, with ablations, alternative alignment recipes, and a probing study, substantiate it. Future work includes longer signals and other modalities.

%% file: 99_acknowledgement.tex
\section*{Acknowledgment}
This research was partially funded by the National Institutes of Health (NIH) under award 1OT2OD038051. The views and conclusions contained in this document are those of the authors and should not be
interpreted as representing the official policies, either expressed or implied, of the NIH.

%% file: tex/5_appendix.tex

This technical appendix provides the full implementation details of
\ourmethod{} that are summarized in the main paper. All values below are the
settings used to produce the reported results. Our code and configuration files
are released for reproducibility.

\section{Datasets and Preprocessing}
We use \mbox{TSFragment-600K} \citep{t2s} with four sources (ETTh1s, ETTm1,
electricity, traffic) at three fragment lengths $L\in\{24,48,96\}$. Each
(source, length) configuration is split $90/10$ into training and held-out test sets. In Stage~1 we replace
Chronos-2's per-instance normalization with a dataset-level global $z$-score
normalization whose mean and standard deviation are estimated on the training
split only, so that absolute level and scale survive into the embedding.

\section{Stage-1: Cross-Modal Alignment}
Table~\ref{tab:stage1} lists the Stage-1 aligner settings. The auxiliary
denoiser is conditioned on the text embedding, trained jointly with the
contrastive objective, and discarded after Stage~1.

\begin{table}[t]
\centering
\begin{minipage}{0.80\textwidth}
\hptable
\begin{tabular}{ll}
\toprule
\textbf{Hyperparameter} & \textbf{Value} \\
\midrule
\multicolumn{2}{l}{\emph{Encoders}}\\
Text encoder & Embedding-Gemma-300M (frozen, LoRA) \\
Series encoder & Chronos-2 (frozen, LoRA) \\
Shared embedding dim $d$ & $512$ \\
Series normalization & dataset-level global $z$-score \\
\addlinespace
\multicolumn{2}{l}{\emph{LoRA}}\\
Rank $r$ & $64$ \\
$\alpha$ & $16$ \\
Dropout & $0.05$ \\
Target modules & $\{q,k,v,o\}$ (both towers) \\
Adapted towers & both (text \& series) \\
\addlinespace
\multicolumn{2}{l}{\emph{Contrastive objective}}\\
Temperature $\tau$ & $0.07$ (fixed) \\
Loss & symmetric InfoNCE $+\ \lambda_{\text{dw}}\,\mathcal{L}_{\text{gen}}$ \\
$\lambda_{\text{dw}}$ (default) & $1.0$ \\
\addlinespace
\multicolumn{2}{l}{\emph{Optimization}}\\
Optimizer & AdamW ($\beta{=}(0.9,0.999)$) \\
Learning rate & $2\times10^{-5}$ (constant, no warmup) \\
Weight decay & $0$ \\
Gradient clipping & $1.0$ \\
Batch size & $256$ \\
Epochs & $100$ \\
\addlinespace
\multicolumn{2}{l}{\emph{Auxiliary denoiser (discarded after Stage~1)}}\\
Backbone & JiT (2-D), hidden $64$, depth $2$, heads $4$ \\
Patch size / image side $m$ & $2$ / $10$ \\
Timestep schedule & logit-normal ($P_{\text{mean}}{=}0.5$, $P_{\text{std}}{=}1.2$) \\
\bottomrule
\end{tabular}
\caption{Stage-1 alignment hyperparameters. In the fine-tuning-method ablation,
FT-Full uses learning rate $1\times10^{-5}$; all other settings are unchanged.}
\label{tab:stage1}
\end{minipage}
\end{table}

\section{Stage-2: Conditional Generation}
Table~\ref{tab:stage2}, Table~\ref{tab:delay}, and Table~\ref{tab:sched} list the Stage-2 generator settings. The generator is
conditioned only on the frozen $512$-d caption embedding and the timestep.

\begin{table}[t]
\centering
\begin{minipage}{0.75\textwidth}
\hptable
\begin{tabular}{ll}
\toprule
\textbf{Hyperparameter} & \textbf{Value} \\
\midrule
\multicolumn{2}{l}{\emph{Backbone (JiT/DiT-2D)}}\\
Normalization / attention & RMSNorm, QK-normalized \\
Positional encoding & 2-D RoPE \\
Feed-forward / conditioning & SwiGLU, adaLN-Zero \\
Hidden size & $128$ \\
Depth (layers) & $4$ \\
Heads (head dim) & $4$ ($32$) \\
Patch size & $2$ \\
Input channels & $1$ \\
Bottleneck dim & $128$ \\
Dropout (attn / proj) & $0.1$ / $0.1$ (middle blocks) \\
Register tokens & $0$ \\
Guidance / condition dropout & none / none \\
\addlinespace
\multicolumn{2}{l}{\emph{Flow matching}}\\
Target & $X$-prediction + velocity matching \\
Timestep schedule & logit-normal ($P_{\text{std}}{=}1.2$) \\
$P_{\text{mean}}$ & per dataset (Table~\ref{tab:sched}) \\
Noise scale $\sigma$ & $1.0$ \\
\addlinespace
\multicolumn{2}{l}{\emph{Sampling}}\\
Solver & Heun (deterministic ODE), $50$ steps \\
EMA decays & $0.9999$ (ema1), $0.999$ (ema2) \\
\addlinespace
\multicolumn{2}{l}{\emph{Optimization}}\\
Optimizer & AdamW ($\beta{=}(0.9,0.999)$) \\
Learning rate & $1\times10^{-4}$ (constant, no warmup) \\
Weight decay & $0$ \\
Gradient clipping & $1.0$ \\
Batch size & $512$ \\
Epochs & $2000$ \\
\bottomrule
\end{tabular}
\caption{Stage-2 generator hyperparameters.}
\label{tab:stage2}
\end{minipage}
\end{table}

\section{Length- and Dataset-Specific Settings}
The delay embedding maps a length-$L$ fragment to a square $m\times m$ image
with column stride $s$; Table~\ref{tab:delay} lists the values per length.
Table~\ref{tab:sched} lists the per-(dataset, length) logit-normal mean
$P_{\text{mean}}$; the standard deviation is $P_{\text{std}}{=}1.2$ throughout.

\begin{table}[!tbp]
\centering
\begin{minipage}[t]{0.48\textwidth}\vspace{0pt}
\centering
\hptable
\adjustbox{max width=\linewidth}{\begin{tabular}{cccc}
\toprule
\textbf{Length }$L$ & \textbf{Image side }$m$ & \textbf{Stride }$s$ & \textbf{Image} \\
\midrule
$24$ & $6$ & $4$ & $6\times6$ \\
$48$ & $8$ & $6$ & $8\times8$ \\
$96$ & $10$ & $10$ & $10\times10$ \\
\bottomrule
\end{tabular}}
\caption{Delay-embedding geometry per fragment length. Traffic at $L{=}24$ uses
$m{=}8$, $s{=}8$ ($8\times8$).}
\label{tab:delay}
\vspace{14pt}

\hptable
\adjustbox{max width=\linewidth}{\begin{tabular}{lccc}
\toprule
\textbf{Dataset} & $L{=}24$ & $L{=}48$ & $L{=}96$ \\
\midrule
ETTh1s & $0.5$ & $0.5$ & $0.5$ \\
ETTm1 & $0.8$ & $0.5$ & $0.0$ \\
electricity & $0.5$ & $0.5$ & $0.5$ \\
traffic & $-0.8$ & $-0.5$ & $-0.8$ \\
\bottomrule
\end{tabular}}
\caption{Per-(dataset, length) logit-normal timestep mean $P_{\text{mean}}$. The
standard deviation is $P_{\text{std}}{=}1.2$ in all settings.}
\label{tab:sched}
\end{minipage}
\hfill
\begin{minipage}[t]{0.48\textwidth}\vspace{0pt}
\centering
\hptable
\adjustbox{max width=\linewidth}{\begin{tabular}{ll}
\toprule
\textbf{Setting} & \textbf{Value} \\
\midrule
Benchmark & \mbox{TSFragment-600K}, 4 sources \\
Series normalization & global min--max to $[0,1]$ \\
Train / val split & $90/10$, split seed $0$ \\
Val-loss seed & $12345$ \\
Gen-eval cadence & every $100$ epochs \\
Gen-eval sample budget & full validation split \\
Metric embedder & frozen reference aligner \\
Metric seed / top-$K$ / bins & $123$ / $5$ / $32$ \\
Tracked weight variants & online, ema1, ema2 \\
Checkpoint selection & lowest FID \\
Data workers & $8$--$16$ \\
\bottomrule
\end{tabular}}
\caption{Settings shared by all five baselines and by \ourmethod{}. Checkpoint
selection and the metric embedder follow the protocol in the Evaluation Protocol
section, so no method is scored in its own embedding space.}
\label{tab:baseline-shared}
\end{minipage}
\end{table}

\section{Baseline Implementations and Hyperparameters}
Every baseline is run inside the same harness as \ourmethod{}: each method's
generator and conditioning path follow its official architecture and released
configuration, but data loading, train/validation splitting, checkpointing and
generation evaluation are shared code, so all methods see identical data and are
scored in the identical embedding space. Table~\ref{tab:baseline-shared} lists
the settings that are common to all five baselines; the per-method settings
follow in Table~\ref{tab:baseline-verbalts}--Table~\ref{tab:baseline-pretrain}.

\subsection{VerbalTS}
VerbalTS \citep{verbalts} is an observation-space DDPM with multi-scale patch
denoising. We keep its released text path exactly: frozen LongCLIP
(\texttt{zer0int/LongCLIP-GmP-ViT-L-14}) per-token features, with no
normalization or adapter inserted between the cached text features and
VerbalTS's own trainable text projector.

\begin{table}[!tbp]
\centering
\begin{minipage}[t]{0.48\textwidth}\vspace{0pt}
\centering
\hptable
\adjustbox{max width=\linewidth}{\begin{tabular}{ll}
\toprule
\textbf{Hyperparameter} & \textbf{Value} \\
\midrule
\multicolumn{2}{l}{\emph{Text path}}\\
Encoder & frozen LongCLIP-GmP-ViT-L-14 \\
Readout & per-token, $T{=}248$, $D{=}768$ \\
Encoder dtype & bfloat16 \\
Pre-projector & $768{\to}512{\to}64$ (LN, LReLU) \\
Text projector stages & $3$ \\
\addlinespace
\multicolumn{2}{l}{\emph{Denoiser}}\\
Variates $n_{\text{var}}$ & $1$ \\
Model width $d$ & $64$ \\
Multi-patch scales $K$ & $3$ (lengths $3,6,12$) \\
Residual blocks / heads & $3$ / $8$ \\
Diffusion-embedding dim & $128$ \\
Variate / time side-emb. & $16$ / $112$ \\
Attention mask & parallel (block-diagonal) \\
Conditioning & adaLN \\
\addlinespace
\multicolumn{2}{l}{\emph{Diffusion}}\\
Steps / schedule & $50$ / quad \\
$\beta$ range & $10^{-4}\to0.5$ \\
Sampler & DDIM, all $50$ steps \\
Prior noise scale & $1.0$ \\
EMA decays & $0.999$ (ema1), $0.9999$ (ema2) \\
\addlinespace
\multicolumn{2}{l}{\emph{Optimization}}\\
Optimizer & AdamW \\
Learning rate / decay & $1\times10^{-3}$ / $0$ \\
Gradient clipping & $1.0$ \\
Batch size / epochs & $512$ / $2000$ \\
\bottomrule
\end{tabular}}
\caption{VerbalTS baseline hyperparameters.}
\label{tab:baseline-verbalts}
\end{minipage}
\hfill
\begin{minipage}[t]{0.48\textwidth}\vspace{0pt}
\centering
\hptable
\adjustbox{max width=\linewidth}{\begin{tabular}{ll}
\toprule
\textbf{Hyperparameter} & \textbf{Value} \\
\midrule
\multicolumn{2}{l}{\emph{Text path}}\\
Encoder & frozen LongCLIP-GmP-ViT-L-14 \\
Readout & pooled CLS \\
Adapter & LayerNorm, hidden $512$ \\
Harness condition dropout & $0$ (disabled) \\
\addlinespace
\multicolumn{2}{l}{\emph{Denoiser}}\\
Input channels & $1$ \\
Prototype dim / count & $32$ / $16$ \\
Latent unit & $1$ \\
Model channels & $32$ \\
Residual blocks per level & $2$ \\
Channel multipliers & $(1,2,4,4)$ \\
Attention resolutions & $(1,2,4)$ \\
Attention heads & $8$ \\
Transformer depth & $1$ \\
Dropout & $0$ \\
Text--prototype fusion & gated add \\
\addlinespace
\multicolumn{2}{l}{\emph{Diffusion}}\\
Steps / schedule & $50$ / quad \\
$\beta$ range & $10^{-4}\to0.5$ \\
Sampler / $\eta$ & DDIM (all steps) / $0.0$ \\
Prototype drop (train) & $0.5$ \\
Guidance scale (sampling) & $1.0$ \\
EMA decays & $0.999$ (ema1), $0.9999$ (ema2) \\
\addlinespace
\multicolumn{2}{l}{\emph{Optimization}}\\
Optimizer & AdamW \\
Learning rate / decay & $1\times10^{-3}$ / $10^{-4}$ \\
Gradient clipping & $1.0$ \\
Batch size / epochs & $256$ / $2000$ \\
\bottomrule
\end{tabular}}
\caption{BRIDGE baseline hyperparameters.}
\label{tab:baseline-bridge}
\end{minipage}
\end{table}

\subsection{BRIDGE}
BRIDGE \citep{bridge} is an observation-space DDPM whose 1-D U-Net is
cross-attended by domain prototypes extracted from an example series, with the
pooled text embedding fused in by a gated conditioning MLP. Classifier-free
guidance is BRIDGE's own mechanism (dropping the prototype context to a
learnable null embedding), so the harness-level condition dropout is disabled to
avoid stacking two guidance mechanisms.

\subsection{T2S}
T2S \citep{t2s} is a two-stage latent method: a length-adaptive VAE is
pretrained first, then a rectified-flow Transformer is trained in its latent
space. Following the released implementation, conditioning uses the $128$-d
\texttt{TextEmbedding} column shipped with \mbox{TSFragment-600K} (OpenAI
embeddings) rather than a text encoder trained here, and the latent geometry is
fixed to the $(64,30)$ tensor the T2S Transformer expects.

\begin{table}[!tbp]
\centering
\begin{minipage}[t]{0.48\textwidth}\vspace{0pt}
\centering
\hptable
\adjustbox{max width=\linewidth}{\begin{tabular}{ll}
\toprule
\textbf{Hyperparameter} & \textbf{Value} \\
\midrule
Text condition & $128$-d dataset TextEmbedding \\
Latent shape & $(64,30)$, from the LA-VAE \\
Generator & rectified-flow Transformer \\
Train-time flow grid & $100$ steps \\
Sampling steps & $10$ (Euler) \\
Guidance scale & $9.0$ \\
Condition dropout (train) & $0.3$ \\
EMA decays & $0.9999$ (ema1), $0.999$ (ema2) \\
Optimizer & AdamW \\
Learning rate / decay & $1\times10^{-4}$ / $0$ \\
Gradient clipping & $1.0$ \\
Batch size / epochs & $256$ / $2000$ \\
\bottomrule
\end{tabular}}
\caption{T2S baseline hyperparameters (stage 2). The stage-1 LA-VAE is in
Table~\ref{tab:baseline-pretrain}.}
\label{tab:baseline-t2s}
\end{minipage}
\hfill
\begin{minipage}[t]{0.48\textwidth}\vspace{0pt}
\centering
\hptable
\adjustbox{max width=\linewidth}{\begin{tabular}{ll}
\toprule
\textbf{Hyperparameter} & \textbf{Value} \\
\midrule
Text condition & $128$-d dataset TextEmbedding \\
Latent shape & $(4, L/2)$, from the VAE \\
U-Net levels / kernel / heads & $3$ / $3$ / $8$ \\
Diffusion steps / schedule & $50$ / quad \\
$\beta$ range & $10^{-4}\to0.5$ \\
Sampler & DDIM \\
Guidance scale & $1.0$ (none) \\
Condition dropout (train) & $0$ \\
EMA decays & $0.9999$ (ema1), $0.999$ (ema2) \\
Optimizer & AdamW \\
Learning rate / decay & $1\times10^{-3}$ / $10^{-4}$ \\
Gradient clipping & $1.0$ \\
Batch size / epochs & $256$ / $2000$ \\
\bottomrule
\end{tabular}}
\caption{DiffuSETS baseline hyperparameters (stage 2). The stage-1 VAE is in
Table~\ref{tab:baseline-pretrain}.}
\label{tab:baseline-diffusets}
\end{minipage}
\end{table}

\subsection{DiffuSETS}
DiffuSETS \citep{diffusets} is a latent DDPM with a U-Net denoiser whose
cross-attention reads the caption embedding. As with T2S, the conditioning
signal is the dataset's $128$-d TextEmbedding column. One VAE is trained per
(source, length) rather than across lengths. The guidance scale is kept at
$1.0$, i.e.\ no classifier-free guidance, which is faithful to the released
setting; condition dropout is correspondingly $0$.

\subsection{Text2Motion}
Text2Motion \citep{text2motion} is adapted from text-to-motion synthesis: a
movement autoencoder is pretrained per (source, length), and a text-conditioned
autoregressive CVAE then generates in its latent space. The optimizer settings
follow the released Comp trainer (plain Adam, gradient clipping $0.5$, no weight
decay); conditioning again uses the dataset's $128$-d TextEmbedding column.

\begin{table}[!tbp]
\centering
\begin{minipage}[t]{0.48\textwidth}\vspace{0pt}
\centering
\hptable
\adjustbox{max width=\linewidth}{\begin{tabular}{ll}
\toprule
\textbf{Hyperparameter} & \textbf{Value} \\
\midrule
Text condition & $128$-d dataset TextEmbedding \\
Text latent dim & $512$ \\
Attention vector dim & $512$ \\
Latent dim $z$ & $128$ \\
Prior / post.\ / dec.\ hidden & $1024$ / $1024$ / $1024$ \\
Prior / post.\ / dec.\ layers & $1$ / $1$ / $1$ \\
Recon.\ weights (series / mov.) & $1.0$ / $1.0$ \\
KL weight & $0.005$ \\
Teacher-forcing ratio & $0.4$ \\
EMA decays & $0.9999$ (ema1), $0.999$ (ema2) \\
Optimizer & Adam \\
Learning rate / decay & $2\times10^{-4}$ / $0$ \\
Gradient clipping & $0.5$ \\
Batch size / epochs & $32$ / $2000$ \\
\bottomrule
\end{tabular}}
\caption{Text2Motion baseline hyperparameters (stage 2). The stage-1 movement
autoencoder is in Table~\ref{tab:baseline-pretrain}.}
\label{tab:baseline-text2motion}
\end{minipage}
\hfill
\begin{minipage}[t]{0.48\textwidth}\vspace{0pt}
\centering
\hptable
\adjustbox{max width=\linewidth}{\begin{tabular}{ll}
\toprule
\textbf{Hyperparameter} & \textbf{Value} \\
\midrule
\multicolumn{2}{l}{\emph{T2S: length-adaptive VAE}}\\
Training lengths & mixed $\{24,48,96\}$ \\
Block hidden / residual layers & $128$ / $2$ \\
Residual hidden / embedding dim & $256$ / $64$ \\
Learning rate & $1\times10^{-3}$ \\
Batch size / epochs / seed & $512$ / $2000$ / $42$ \\
\addlinespace
\multicolumn{2}{l}{\emph{DiffuSETS: convolutional VAE}}\\
Training lengths & one model per length \\
Downsample factor & $2$ \\
Latent / base channels & $4$ / $128$ \\
KL weight (constant) & $0.01$ \\
Optimizer & AdamW \\
Learning rate / decay & $1\times10^{-3}$ / $10^{-4}$ \\
Gradient clipping & $1.0$ \\
Batch size / epochs / seed & $256$ / $1000$ / $42$ \\
\addlinespace
\multicolumn{2}{l}{\emph{Text2Motion: movement autoencoder}}\\
Training lengths & one model per length \\
Movement latent dim & $512$ \\
Enc.\ / dec.\ hidden & $512$ / $512$ \\
Sparsity / smoothness weights & $0.001$ / $0.001$ \\
Optimizer & Adam (no clipping) \\
Learning rate / decay & $1\times10^{-4}$ / $0$ \\
Batch size / epochs / seed & $128$ / $270$ / $42$ \\
\bottomrule
\end{tabular}}
\caption{Stage-1 autoencoder pretraining for the two-stage baselines. Each
stage-1 model reuses its stage-2 validation ratio ($0.1$) and split seed ($0$).}
\label{tab:baseline-pretrain}
\end{minipage}
\end{table}

\subsection{Stage-1 Pretraining for the Two-Stage Baselines}
T2S, DiffuSETS and Text2Motion each require a pretrained autoencoder before
their generator can be trained. In all three cases the stage-1 model uses the
same validation ratio and split seed as its stage-2 generator, so the
autoencoder never sees a held-out fragment. Table~\ref{tab:baseline-pretrain}
lists the settings.

\section{Evaluation Protocol}

\subsection{The Evaluation Aligner}
Fidelity is measured by FID and caption adherence by CTTP ($\uparrow$) and JFTSD
($\downarrow$), computed with our reimplementation of the ConTSG-Bench
metrics~\citep{contsg}. All three live in the embedding space of a single
text--time-series dual encoder (the evaluation aligner) that exists only for
measurement. FID is the Fr\'echet distance between the embeddings of real and
generated fragments, CTTP the mean cosine similarity between a caption embedding
and the embedding of the fragment generated from that caption, and JFTSD the
Fr\'echet distance over the concatenated [series, text] embeddings.

Because \ourmethod{} is itself a text--time-series alignment, an evaluator built
from the same components might favor it by construction. We therefore build the
evaluation aligner from disjoint parts. Its text tower is
BGE-large-en-v1.5~\citep{bge}, a different model family from the
Embedding-Gemma-300M that \ourmethod{} adapts; its series tower is a
convolutional encoder trained from scratch, not initialized from Chronos-2 or
any other time-series foundation model. No weights are shared with either
\ourmethod{} tower in either direction. We do not claim that the two text
encoders were pretrained on disjoint web corpora, since neither model's pretraining
data is public at that granularity; we claim only that the evaluator reuses no
parameters from the models it scores.

One evaluation aligner is fit per (source, length) configuration on that
configuration's training split, under exactly the split used to train every
generator it scores ($10\%$ held out, split seed $0$); we verified that the
held-out fragments on which metrics are computed do not occur in the aligner's
own training data. It is then frozen and reused unchanged for every method,
ablation, and figure in this paper, so all reported numbers are commensurable
and no model is scored in a space it helped produce.

\subsection{Checkpoint Selection and Aggregation}
Generation metrics are
evaluated every $100$ epochs for all three model variants (online, ema1, ema2).
For each run we select the single checkpoint (epoch $\times$ variant) with the
lowest FID on the full held-out split, then re-sample that checkpoint $3$ times
with independent sampling seeds ($100$/$200$/$300$) and report mean $\pm$ sample
standard deviation over those $3$ draws. Both stages are trained once with a
fixed seed ($0$), so the reported variation is sampling variance: it quantifies
how much a reported number moves when the same trained model is re-sampled, and
does not capture run-to-run training variance. We state this explicitly because
the comparisons in Table~1 of the main paper turn on small absolute differences;
Appendix~\ref{app:significance} quantifies that variance and tests the resulting
ordering.

\section{Variability and Significance Testing}
\label{app:significance}

Table~1 of the main paper reports means only, and several columns are decided by
differences in the third decimal. This appendix reports the corresponding
dispersion and tests whether the resulting ordering is statistically supported.

\subsection{Per-Cell Sampling Variability}
Table~\ref{tab:main-std} gives the sample standard deviation over the $3$
sampling seeds for every one of the $216$ cells of Table~1
($6$ methods $\times$ $4$ datasets $\times$ $3$ lengths $\times$ $3$ metrics).
Dispersion is small relative to the reported gaps: the largest standard
deviation anywhere is $0.0165$ (DiffuSETS, FID on ETTh1s-$96$), the largest for
\ourmethod{} is $0.0033$, $98.1\%$ of all cells are below $0.01$, and the median
cell is $0.0011$ (FID), $0.0026$ (CTTP), $0.0024$ (JFTSD). As a per-cell check
against the strongest baseline we ran a Welch $t$-test (\ourmethod{} vs.\
VerbalTS, $n=3$ per side) in each of the $36$ dataset $\times$ length $\times$
metric cells and corrected the $36$ $p$-values with Holm's procedure:
$29$ of the $30$ cells we win remain significant at $\alpha=0.05$, and across
those winning cells the median gap is $34\times$ the pooled standard deviation.
The three cells that are not resolved are traffic-$48$ JFTSD (a win) and the two
FID cells of ETTh1s-$96$ and ETTm1-$96$ (both losses), where the two methods are
within one standard deviation of each other. As noted above, this variance is
sampling-only; it bounds the noise in each reported number but not the
variability of retraining.

\subsection{Wilcoxon Signed-Rank Tests Against the Baselines}

\begin{wraptable}{r}{\aaaicolwidth}
\centering
{\small
\setlength{\tabcolsep}{3pt}
\begin{tabular}{lcccccc}
\toprule
 & \multicolumn{2}{c}{\textbf{FID}} & \multicolumn{2}{c}{\textbf{CTTP}} & \multicolumn{2}{c}{\textbf{JFTSD}} \\
\cmidrule(lr){2-3} \cmidrule(lr){4-5} \cmidrule(lr){6-7}
\textbf{Ours vs.} & W--L & $p_{\text{Holm}}$ & W--L & $p_{\text{Holm}}$ & W--L & $p_{\text{Holm}}$ \\
\midrule
VerbalTS    & 7--5  & $0.133$ & 12--0 & $0.0012$ & 11--1 & $0.0012$ \\
Text2Motion & 12--0 & $0.0012$ & 12--0 & $0.0012$ & 12--0 & $0.0012$ \\
BRIDGE      & 11--1 & $0.0068$ & 12--0 & $0.0012$ & 12--0 & $0.0012$ \\
DiffuSETS   & 12--0 & $0.0012$ & 12--0 & $0.0012$ & 12--0 & $0.0012$ \\
T2S         & 12--0 & $0.0012$ & 12--0 & $0.0012$ & 12--0 & $0.0012$ \\
\bottomrule
\end{tabular}}
\caption{Pairwise one-sided Wilcoxon signed-rank tests of \ourmethod{} against
each baseline, run separately per metric over the $12$ settings (four datasets
$\times$ three lengths), with Holm correction over the five comparisons within a
metric. W--L counts the settings won and lost. The only comparison that fails to
reach significance is FID against VerbalTS, i.e.\ the two are indistinguishable
on fidelity.}
\label{tab:wilcoxon}
\end{wraptable}

Following the standard protocol for comparing methods over multiple problems
\citep{demsar}, we test \ourmethod{} against each baseline with a one-sided
Wilcoxon signed-rank test, Holm-corrected over the five comparisons. We run
these per metric, so that the $12$ paired observations entering a test (four
datasets $\times$ three lengths) are all on the same scale; a signed-rank test
that mixed FID, CTTP, and JFTSD in one sample would weight the metrics by their
arbitrary numeric ranges. Table~\ref{tab:wilcoxon} reports one-sided
$p$-values.

The outcome is uniform against four of the five baselines: \ourmethod{} wins
$12/12$ settings on essentially every metric and every Holm-corrected $p$ is
below $0.01$. Against VerbalTS the picture is split. On both caption-adherence
metrics the advantage is significant (CTTP $12/12$, JFTSD $11/12$, both
$p=1.2\times10^{-3}$), whereas on FID it is not ($7/12$ wins, $p=0.13$): the two
methods are statistically indistinguishable on fidelity. We therefore do not
claim a fidelity advantage over VerbalTS, only parity, and locate our gain on
caption adherence.

For completeness we also ran the test in the form suggested by the layout of
Table~1, i.e.\ on the $12$ columns of a single length taken together, converting
each column to the relative improvement of \ourmethod{} over VerbalTS so that
the three metric scales are commensurate. This gives $p=7\times10^{-4}$ at
$L=24$ ($11/12$ columns won, median relative gain $30.3\%$), $p=7\times10^{-4}$
at $L=48$ ($11/12$, $37.3\%$), $p=0.12$ at $L=96$ ($8/12$, $12.8\%$), and
$p=3.7\times10^{-6}$ on all $36$ columns pooled ($30/36$, $28.6\%$).

A per-length test of this kind pools the three metrics, so it inherits the FID
parity reported above, and that is what the $L=96$ value reflects rather than
any weakening of caption adherence. Reading the same results one (dataset,
length) pair at a time confirms this. A signed-rank test cannot be run at that
granularity, because each pair contributes only three columns, so the smallest
attainable one-sided $p$ is $0.125$; we therefore use the per-cell Welch
tests of the previous subsection. \ourmethod{} wins \emph{both}
caption-adherence metrics in $11$ of the $12$ pairs, and all three metrics in
$7$ of them; five of the six lost cells are FID, the sixth being JFTSD on
traffic-$96$, the only pair in which an adherence metric goes against us. By
metric, \ourmethod{} wins CTTP on $4/4$ domains at \emph{every} length and JFTSD
on $4/4$, $4/4$ and $3/4$ domains at $L=24/48/96$, whereas FID goes $3/4$, $3/4$
and $1/4$. The caption-adherence advantage therefore holds at $L=96$ much as it
does at the two shorter lengths; what narrows at $L=96$ is fidelity, on which we
claim only parity in the first place.

\begin{table}[t]
\centering
{\small
\setlength{\tabcolsep}{1mm}
\begin{tabular}{lcccccccccccc}
\toprule
\textbf{Method} & \multicolumn{3}{c}{\textbf{ETTh1s}} & \multicolumn{3}{c}{\textbf{ETTm1}} & \multicolumn{3}{c}{\textbf{Electricity}} & \multicolumn{3}{c}{\textbf{Traffic}} \\
\cmidrule(lr){2-4} \cmidrule(lr){5-7} \cmidrule(lr){8-10} \cmidrule(lr){11-13}
 & FID & CTTP & JFTSD & FID & CTTP & JFTSD & FID & CTTP & JFTSD & FID & CTTP & JFTSD \\
\midrule
\multicolumn{13}{l}{\emph{Length }$L=24$} \\
\midrule
DiffuSETS   & 3.4 & 4.4 & 5.4 & 1.6 & 1.1 & 2.3 & 3.3 & 4.4 & 3.6 & 1.7 & 1.0 & 2.6 \\
T2S         & 1.0 & 6.6 & 6.3 & 1.9 & 0.7 & 3.2 & 2.6 & 4.7 & 2.8 & 6.1 & 1.3 & 4.3 \\
BRIDGE      & 0.5 & 8.2 & 8.0 & 1.5 & 2.3 & 2.6 & 1.9 & 3.2 & 3.8 & 9.1 & 3.7 & 7.3 \\
Text2Motion & 0.9 & 2.1 & 1.2 & 0.6 & 3.0 & 2.4 & 1.5 & 3.0 & 2.1 & 0.4 & 3.9 & 1.7 \\
VerbalTS    & 0.4 & 5.4 & 3.9 & 0.1 & 0.3 & 0.5 & 0.6 & 2.6 & 1.4 & 1.5 & 4.1 & 2.6 \\
\textbf{Ours} & 0.2 & 1.4 & 0.4 & 0.1 & 0.5 & 0.1 & 0.2 & 2.3 & 0.5 & 0.2 & 0.3 & 0.2 \\
\midrule
\multicolumn{13}{l}{\emph{Length }$L=48$} \\
\midrule
DiffuSETS   & 1.3 & 8.8 & 7.7 & 1.2 & 1.0 & 2.0 & 1.6 & 2.1 & 2.5 & 0.9 & 3.2 & 4.1 \\
T2S         & 0.6 & 3.7 & 3.3 & 1.9 & 0.2 & 2.2 & 1.7 & 2.8 & 2.9 & 0.7 & 3.5 & 2.5 \\
BRIDGE      & 1.0 & 6.5 & 6.5 & 1.4 & 1.7 & 1.6 & 1.3 & 2.6 & 2.9 & 4.7 & 6.3 & 7.1 \\
Text2Motion & 0.7 & 4.2 & 3.6 & 0.8 & 2.4 & 2.5 & 1.0 & 1.5 & 1.7 & 1.5 & 2.2 & 1.8 \\
VerbalTS    & 0.9 & 3.8 & 5.5 & 0.2 & 2.4 & 0.7 & 0.8 & 2.8 & 1.4 & 1.4 & 3.0 & 1.9 \\
\textbf{Ours} & 0.1 & 3.3 & 1.4 & 0.1 & 0.3 & 0.2 & 1.0 & 1.3 & 0.8 & 0.3 & 1.2 & 0.6 \\
\midrule
\multicolumn{13}{l}{\emph{Length }$L=96$} \\
\midrule
DiffuSETS   & 16.5 & 2.8 & 10.2 & 2.1 & 1.1 & 2.4 & 6.7 & 3.7 & 8.1 & 12.8 & 2.0 & 13.0 \\
T2S         & 0.6 & 7.5 & 8.8 & 0.5 & 1.4 & 1.4 & 1.8 & 1.3 & 2.2 & 8.2 & 1.1 & 5.2 \\
BRIDGE      & 3.6 & 5.0 & 4.0 & 0.9 & 1.9 & 0.3 & 1.9 & 4.1 & 1.7 & 5.0 & 3.8 & 1.7 \\
Text2Motion & 3.0 & 4.1 & 6.3 & 1.5 & 0.7 & 1.4 & 1.5 & 4.4 & 3.1 & 6.7 & 5.5 & 6.2 \\
VerbalTS    & 1.3 & 8.1 & 5.0 & 0.3 & 1.7 & 0.6 & 0.8 & 1.7 & 1.6 & 0.5 & 3.0 & 0.3 \\
\textbf{Ours} & 0.7 & 1.5 & 0.8 & 0.7 & 0.7 & 0.5 & 0.4 & 0.7 & 0.3 & 1.1 & 1.8 & 1.3 \\
\bottomrule
\end{tabular}}
\caption{Sample standard deviation over the $3$ sampling seeds for every cell of
Table~1 in the main paper. \textbf{All values are $\times10^{-3}$}, so an entry
of $3.4$ means $0.0034$. The largest entry in the whole table is $16.5$
($0.0165$; DiffuSETS, FID on ETTh1s-$96$) and the largest for \ourmethod{} is
$3.3$ ($0.0033$).}
\label{tab:main-std}
\end{table}

\section{Morphological Diversity of the Benchmark Sources}
\label{app:diversity}

Traffic is where \ourmethod{}'s margin is thinnest: three of the six cells it
loses to VerbalTS sit there, and traffic-$96$ is the only (dataset, length) pair
in which a caption-adherence metric goes against us. This appendix characterizes
why, by measuring how many distinct \emph{shapes} each source actually contains.

\paragraph{Protocol}
Sources differ by orders of magnitude in absolute scale, so we first
$z$-normalize every fragment individually; this removes level and amplitude and
leaves only morphology. On $2000$ fragments per (source, length) we then compute
the PCA spectrum of the shape population and a $k{=}6$ KMeans clustering, and
report: the share of shape variance in the first two components
(PCA$_{2}$); the number of components needed to reach $90\%$ (PC$_{90}$); and
the participation ratio $(\sum_i\lambda_i)^2/\sum_i\lambda_i^2$ of the spectrum,
an effective number of shape dimensions (eff.\ rank). A source with few distinct
shapes has high PCA$_2$ and low PC$_{90}$ and eff.\ rank.

\begin{figure}[t]
\centering
\includegraphics[width=\textwidth]{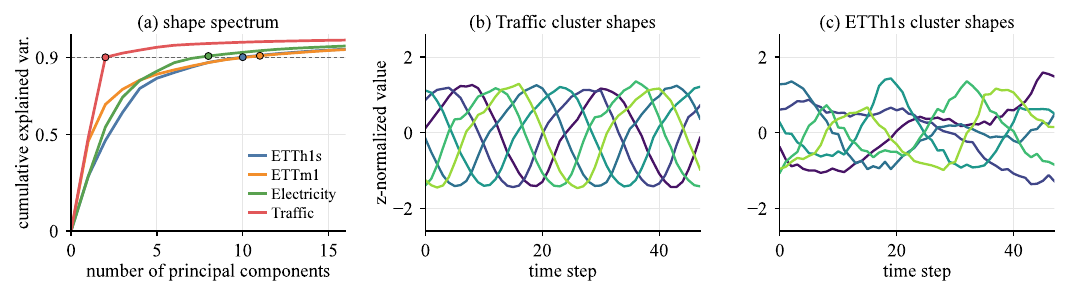}
\caption{Morphological diversity at $L=48$. (a) Cumulative PCA explained
variance of the $z$-normalized fragments; markers sit where each source reaches
$90\%$ (dashed line). Traffic saturates after two components while the others
need eight to eleven. (b, c) The six KMeans centroid shapes of Traffic and of
ETTh1s on identical axes: Traffic's clusters are one daily template at six
different phases, whereas ETTh1s's are distinct morphologies.}
\label{fig:diversity}
\end{figure}

\paragraph{Traffic is an extreme outlier}
Table~\ref{tab:diversity} shows the result, and the ordering is the same at all
three lengths. Traffic needs only $2$--$4$ principal components to explain $90\%$
of its shape variance, against $6$--$11$ for the other three sources, and its
effective rank ($2.3$--$2.6$) is roughly half of theirs. Its KMeans clusters are
also the tightest and best separated (silhouette $0.35$--$0.43$ versus
$0.15$--$0.25$ elsewhere), and random fragment pairs are far more correlated
(mean $|r|$ $0.55$--$0.59$ versus $0.31$--$0.45$).
Figure~\ref{fig:diversity} makes the reason visible: the six Traffic cluster
centroids are the same daily template read at six different phases, whereas the
ETTh1s centroids are genuinely different morphologies. Traffic is, in effect, a
one-waveform benchmark.

\paragraph{How this relates to our results, and what it does not show}
A source with essentially one shape family leaves a caption little morphology to
specify, so there is less for a text--series alignment to exploit; a narrow real
distribution also makes fidelity easy to reach by generating near the mode.
Consistent with that, averaged over the three lengths our relative gain over
VerbalTS on traffic is $+21.7\%$ on CTTP, $+0.1\%$ on JFTSD and $-11.1\%$ on FID,
against $+30.8\%$, $+54.5\%$ and $+20.7\%$ on the other three sources.

We deliberately stop short of a causal claim. Across the $12$ (source, length)
configurations the rank correlation between effective shape rank and our
relative gain is not significant for any metric (Spearman $\rho=+0.01$ for CTTP,
$+0.29$ for JFTSD, $+0.34$ for FID; all $p>0.25$), which four sources are far too
few to resolve. We therefore report the diversity measurement as a
characterization of the benchmark, in that traffic is quantitatively degenerate
in morphology at every length, and as the most plausible reading of where our
method has the least room to help, not as a demonstrated mechanism.

\begin{table}[t]
\centering
{\small
\setlength{\tabcolsep}{6pt}
\begin{tabular}{lccccccccc}
\toprule
 & \multicolumn{3}{c}{$L=24$} & \multicolumn{3}{c}{$L=48$} & \multicolumn{3}{c}{$L=96$} \\
\cmidrule(lr){2-4} \cmidrule(lr){5-7} \cmidrule(lr){8-10}
\textbf{Source} & PCA$_2$ & PC$_{90}$ & eff. & PCA$_2$ & PC$_{90}$ & eff. & PCA$_2$ & PC$_{90}$ & eff. \\
\midrule
ETTh1s      & $0.62$ & $8$ & $4.6$ & $0.47$ & $10$ & $6.4$ & $0.56$ & $9$  & $4.7$ \\
ETTm1       & $0.64$ & $9$ & $3.6$ & $0.66$ & $11$ & $3.8$ & $0.62$ & $11$ & $4.7$ \\
Electricity & $0.71$ & $6$ & $3.8$ & $0.54$ & $8$  & $5.5$ & $0.44$ & $11$ & $7.2$ \\
\midrule
Traffic     & $\mathbf{0.93}$ & $\mathbf{2}$ & $\mathbf{2.3}$ & $\mathbf{0.90}$ & $\mathbf{2}$ & $\mathbf{2.5}$ & $\mathbf{0.87}$ & $\mathbf{4}$ & $\mathbf{2.6}$ \\
\bottomrule
\end{tabular}}
\caption{Shape diversity of the four sources, measured on $2000$ per-series
$z$-normalized fragments per configuration. PCA$_2$ is the share of shape
variance in the first two principal components, PC$_{90}$ the number of
components reaching $90\%$, and eff.\ the participation ratio of the spectrum.
Traffic (bold) needs the fewest dimensions at every length, i.e.\ it contains
the fewest distinct shapes.}
\label{tab:diversity}
\end{table}

\section{Sweeps}
We report the following sweeps in the main paper: the auxiliary-loss weight
$\lambda_{\text{dw}}\in\{0,1,2,4,8\}$; the fine-tuning method
$\in\{\text{LoRA},\text{FT-Full},\text{FT-Head}\}$; the adapted tower
$\in\{\text{text-only},\text{series-only},\text{both}\}$; and the series
normalization $\in\{\text{global }z\text{-score},\text{ per-sample}\}$. In every
sweep, the cell that coincides with our default configuration reports the same
numbers as the corresponding entry of Table~1 in the main paper.

\section{Software and Hardware}
All experiments run on a single NVIDIA RTX~PRO 6000 Blackwell GPU. Our
implementation uses Python~3.11, PyTorch~2.11.0, HuggingFace Transformers~4.57.6,
PEFT~0.18.1, chronos-forecasting~2.2.2, Accelerate~1.12.0, and NumPy~2.1.3.